\documentclass[pdflatex,sn-mathphys-num]{sn-jnl} 
\usepackage{graphicx}%
\usepackage{multirow}%
\usepackage{amsmath,amssymb,amsfonts}%
\usepackage{amsthm}%
\usepackage{mathrsfs}%
\usepackage[title]{appendix}%
\usepackage{xcolor}%
\usepackage{textcomp}%
\usepackage{manyfoot}%
\usepackage{booktabs}%
\usepackage{algorithm}
\usepackage{algorithmicx}
\usepackage{algpseudocode}
\usepackage{listings}%
\usepackage{subcaption}
\usepackage{pgfplots}
\usepackage{pgfplotstable}
\usepgfplotslibrary{patchplots}
\pgfplotsset{compat=1.18}
\theoremstyle{thmstyleone}%

\theoremstyle{thmstyletwo}%
\theoremstyle{thmstylethree}%
\begin{document}
	
	\title[Article Title]{The Anatomy of Implicit Bias: Information Allocation in Neural Network Training}
	
	\author[1]{\fnm{Zhang} \sur{Gongyue}}
	\author[1]{\fnm{Wang} \sur{Zhiyong}}
	\author[1]{\fnm{Liu} \sur{Donghan}}
	\author[1]{\fnm{Ren} \sur{Weihong}}
	\author[1]{\fnm{Sheng} \sur{Yixuan}}
	\author*[1]{\fnm{Liu} \sur{Honghai}}\email{honghai.liu@icloud.com}
	
	
	
	\affil[1]{\orgdiv{State Key Laboratory of Robotics and Systems}, \orgname{Harbin Institute of Technology Shenzhen}, \orgaddress{\city{Shenzhen}, \postcode{518055}, \country{China}}}
	
	\abstract{
		Implicit bias is usually explained as the preference of an optimization process for certain final solutions and their geometry. This view helps explain where a model finally stops. It gives less direct explanation of how this bias is formed during training. This paper proposes a training-time information allocation view. Under this view, optimization forms a writing pattern for error signals across parameter paths, coordinate channels, and sample regions.
		
		This paper builds a set of observable allocation diagnostics. These diagnostics include gradient demand, actual update injection, coordinate gain induced by exponential moving averages, channel-level update ratios, and sample-wise loss distributions. To separate training progress from internal allocation, this paper introduces a collapse--persistence analysis. Under matched training loss, if external loss statistics collapse but internal allocation ratios remain separated, then the factor changes the internal allocation of the training signal. Controlled experiments show that the effect of the learning rate largely collapses after progress matching. Thus, the learning rate is closer to a progress-dominant source. In contrast, the preconditioning exponent \(p\) keeps clear channel-gain and update-ratio signatures after progress matching. This indicates that \(p\) directly changes coordinate allocation induced by EMA preconditioning. Batch size, optimizer memory, data statistics, model width, activation function, and scheduling also show different allocation signatures. These results suggest that training-time implicit bias has multiple sources.
		
		This paper further verifies these observations on standard neural network tasks. On CIFAR-100 with ResNet-18, changing \(p\) systematically affects the weight--bias gradient structure, sample loss quantiles, and the identity of hard samples. It also shows a median--tail trade-off. A larger \(p\) reduces the loss of most samples more strongly, but it can increase the high-loss tail. On facial expression recognition tasks, dynamic information-allocation schedules improve convergence behavior and final performance. This shows that allocation-aware optimization can be useful in real tasks.
		
		Overall, this paper extends the analysis of implicit bias from final-solution geometry to training-time signal allocation. The main claim is that implicit bias is not only reflected by the final solution. It is also reflected by which parameter paths, coordinate channels, and sample regions receive the error signal first and more strongly during training. Based on this view, this paper places different training factors into a unified information-allocation diagnostic framework. The framework gives a mechanism-level explanation of training-time implicit bias. It also provides a basis for future optimization methods that control training progress and signal allocation separately.
	}
	
	
	\maketitle
	
	\section{Introduction}\label{sec1}
	
	Before discussing neural network optimization, we start with a simple analogy from flight. A gliding seagull and an artificial aircraft can both adjust their posture to correct a flight path and approach a target. However, they rely on different degrees of freedom. A seagull has rich and continuous body control. It has lower requirements on external initial conditions such as altitude. An aircraft depends more strongly on external conditions when it optimizes its flight path. Thus, even for the same task of flying toward a target, different systems allocate path errors to different correction paths.
	
	Neural networks have a similar difference. Models with different sizes and architectures do not only differ in the number of parameters. They also differ in the degrees of freedom used to absorb training error. Wider channels, deeper layers, and richer module combinations provide more input-dependent representation paths. Normalization affine parameters, biases, classifier heads, and other calibration parameters provide scale, shift, and local boundary correction paths. Therefore, the same error signal is not written into all parameters in the same way after backpropagation. Model architecture determines which paths can absorb error. Data statistics determine which signals can be observed. The training mechanism determines the ratio and order in which these signals are written into the model.
	
	For this reason, the training process should not be viewed only as a process of loss reduction. The learning rate mainly controls the global speed of training progress. Adam-like adaptive preconditioning changes the relative gain of different coordinates through the exponential moving average of second moments. Weight decay adds shrinkage pressure on parameters. Batch size changes the stochastic observation of sparse signals in mini-batches. Optimizer memory and training schedules further change how these states persist and appear over time. Together, these factors determine how error signals are written into parameter paths, coordinate channels, and sample regions. Thus, training does not only reduce loss. It also forms an observable information-allocation structure.
	
	This point also reveals an often overlooked fact. Implicit bias is not only present in the final solution space. It is also formed during training. Different first-order optimizers, learning rates, batch sizes, regularization methods, model architectures, and data statistics can change training trajectories, representation structures, and final generalization behavior \cite{pmlr-v235-cattaneo24a,wilson2017marginal,zhang2024implicit,vasudeva2026rich}. In standard training practice, these factors are often discussed separately. The optimizer is treated as a numerical update rule. The learning rate is treated as step-size control. Weight decay is treated as regularization. Batch size is treated as an efficiency or noise parameter. Model architecture and data statistics are analyzed as separate problems. This leaves a more basic question open. Can these heterogeneous factors be compared through a common training-time observation layer? How do they change the flow of training signals?
	
	Existing studies have discussed implicit bias in optimization and training from many angles. SGD \cite{robbins1951stochastic}, Momentum \cite{polyak1964some}, RMSProp \cite{geoffrey2012rmsprop}, Adam \cite{kingma2014adam}, and AdamW \cite{loshchilov2017decoupled} have different update directions, noise properties, and local geometric preferences. Related work often explains these differences through convergence speed, gradient noise, sharpness, flatness, Hessian spectra, or geometry of the solution space \cite{zaheer2018adaptive,li2018visualizing}. These analyses show that the training mechanism affects the region reached by the model. However, they less directly answer a training-time question. After backpropagation, how does the same error signal enter different parameter paths, coordinate channels, and sample regions? How do different training factors change these writing patterns and further shape later gradient statistics and generalization behavior?
	
	This paper starts from the training-time update process and proposes information allocation as a common observation layer. In this paper, information allocation means the relative writing pattern formed by learning signals across parameter paths, coordinate channels, optimizer states, and sample regions. In other words, we do not only ask where the model finally stops. We ask which parameter paths more easily absorb the current error during training, which coordinate channels obtain larger effective gains, which sample regions are fitted first, and how these differences further change later gradient statistics and optimization trajectories.
	
	This view is complementary to the traditional geometric view. Analyses based on sharpness, flatness, margin, and Hessian statistics mainly describe the geometry of the final solution. Information allocation instead focuses on how this geometry is formed during training. A training factor may not directly change a final geometric metric. However, it may continuously change the writing path of training signals. For example, the learning rate changes the global speed of training progress and trajectory feedback. Batch size changes the stochastic observation of sparse signals in mini-batches. Weight decay adds parameter shrinkage pressure. Momentum and second-moment memory change how historical gradients enter the current update. These factors have different mathematical forms, but they can all leave observable allocation signatures during training.
	
	To characterize this allocation structure, this paper constructs four types of diagnostics. The first is gradient demand. It describes the gradient received by different parameter paths or channels. The second is update injection. It describes the actual update written by the optimizer into these paths or channels. The third is optimizer-state gain. It describes the effective coordinate gain induced by EMA second moments, momentum memory, or gain floors. The fourth is sample-level loss distribution. It describes the fitting order and tail distribution of different sample regions during training. By observing these quantities together, we can separate whether a training factor changes where the model is trained to, or whether it changes the internal path by which the model reaches that state.
	
	Among these factors, the continuous preconditioning exponent \(p\) gives a clean analytical interface \cite{chen2020closing}. In Adam-like updates, the second-moment estimate usually enters the denominator through a square root. More generally, the update can be written as
	\begin{equation}
		\Delta\theta_{t,i}
		=
		-\eta_t
		\frac{\hat m_{t,i}}
		{(\hat v_{t,i}+\epsilon)^p},
		\label{eq:intro_p_update}
	\end{equation}
	where \(i\) denotes a parameter coordinate, \(\eta_t\) is the learning rate, and \(\hat m_t\) and \(\hat v_t\) are the first-moment and second-moment estimates. When \(p=0\), the second moment does not explicitly scale the coordinates. The update is close to a momentum-like form. When \(p=1/2\), the update is close to an Adam-like form. Intermediate values correspond to different strengths of diagonal preconditioning. Therefore, \(p\) is not the only source of implicit bias studied in this paper. It is a continuous probe for EMA-dependent coordinate allocation.
	
	The effects of \(p\) and the learning rate must be separated. With fixed gradients and fixed optimizer states, the learning rate mainly applies a global scale to the current update. In contrast, \(p\) changes the relative gain between coordinates through \((\hat v_t+\epsilon)^{-p}\). Thus, \(p\) can directly change coordinate-level and channel-level allocation in the current step. In nonlinear deep networks, a larger learning rate can also change later allocation states by changing the next parameter position, residual structure, and gradient statistics. However, this effect comes from multi-step trajectory feedback. It does not come from a direct change in the relative preconditioning gain in the current step. Therefore, the learning rate is closer to progress-dominant control, while \(p\) is closer to coordinate-gain control. They are coupled during multi-step training, but they control different objects.
	
	To avoid confusing training progress with internal allocation, this paper further introduces a collapse--persistence analysis. Fixed-epoch comparison mixes two questions. One is where the model has been trained to. The other is which internal path the model used to reach that state. We therefore compare different settings under matched training loss. If external loss statistics collapse after matching training progress, while internal allocation ratios remain separated, then the factor changes the internal allocation structure of the training signal. In contrast, if both external metrics and internal metrics collapse under matched-progress comparison, then the factor is closer to a global progress effect. This analysis allows learning rate, \(p\), batch size, optimizer memory, data statistics, model width, activation function, and schedule to be compared under the same diagnostic protocol.
	
	This paper also uses weight and bias-like parameters as a minimal projection of parameter-path allocation. Weight parameters usually participate in input-dependent transformations. Explicit biases, normalization affine parameters, classifier heads, and other similar parameters often participate in shift, scale, and local calibration. These groups do not cover all information-allocation phenomena. However, they provide a parameter-path partition that can be recorded and compared in modern networks. By recording gradient norms, update norms, EMA gains, and relative ratios for weight-like and bias-like parameters, we can observe how different training factors change the writing state of error signals between representation paths and calibration paths.
	
	The experiments follow this framework. First, controlled synthetic tasks are built with dense informative channels, sparse informative channels, and sparse noisy channels. We then compare the allocation signatures of learning rate, \(p\), batch size, optimizer memory, \(\epsilon\), data statistics, model width, activation function, and schedule. Second, on CIFAR-100 with ResNet-18, we study how \(p\) affects the weight--bias gradient structure, sample loss quantiles, and hard-sample migration. This checks the projection of the source map in standard visual training. Finally, on facial expression recognition tasks, we test whether dynamic information-allocation schedules affect convergence and final performance. The goal of this paper is not to claim a single best accuracy. The goal is to understand how training factors change the signal-writing structure during training.
	
	\begin{figure}[t]
		\centering
		\includegraphics[width=\linewidth]{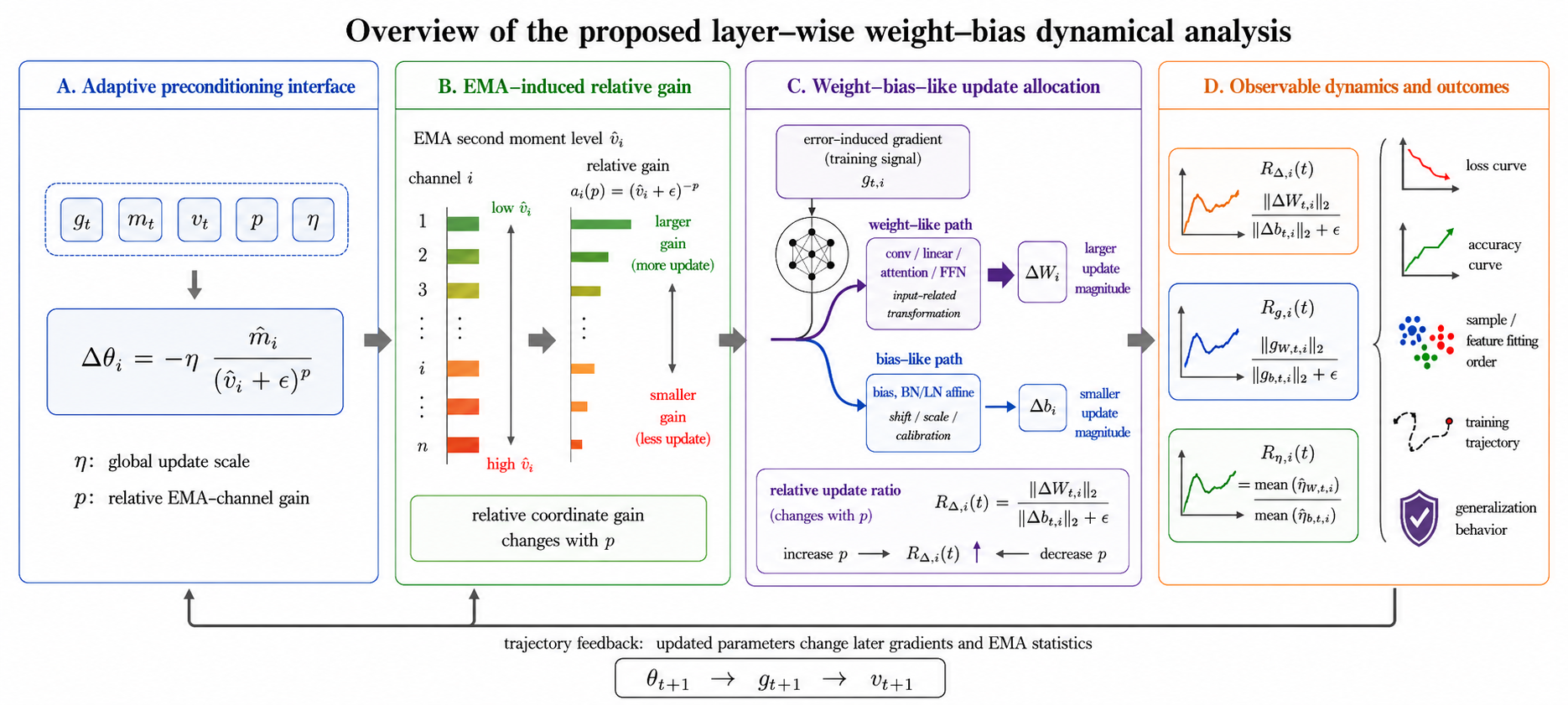}
		\caption{
			Overview of the proposed information-allocation framework.
			Training-time implicit bias is observed through how learning signals are distributed across parameter paths, coordinate channels, optimizer-state gains, and sample regions.
			Different sources act through different signatures: learning rate mainly controls global progress and trajectory feedback; the exponent $p$ controls EMA-dependent coordinate gain; weight decay imposes shrinkage pressure; batch size changes stochastic observation of sparse signals; optimizer memories and gain floors reshape numerator and denominator states; data statistics, model width, activation functions, and schedules further determine available signal paths and their temporal ordering.
			The framework records gradient demand, update injection, gain ratios, channel-level update ratios, and sample-level loss distributions to compare these heterogeneous sources within a common diagnostic layer.
		}
		\label{fig:overview}
	\end{figure}
	
	The main contributions of this paper are as follows.
	
	First, this paper proposes an information-allocation view. It extends the analysis of implicit bias from final-solution geometry to the signal-writing structure during training. This view jointly observes gradient demand, update injection, optimizer-state gain, channel-level update ratio, and sample-level loss distribution. These diagnostics describe how training signals enter different parameter paths, coordinate channels, and sample regions.
	
	Second, this paper builds a source map for training-time implicit bias. It places heterogeneous factors, including learning rate, the preconditioning exponent \(p\), weight decay, batch size, optimizer memory, \(\epsilon\), data statistics, model width, activation function, and schedule, into a common allocation diagnostic framework. This map separates progress-dominant sources, coordinate-gain sources, shrinkage sources, stochastic-observation sources, memory sources, data-statistical sources, and structure-induced sources.
	
	Third, this paper proposes collapse--persistence analysis to separate training progress from internal allocation. By comparing external loss statistics and internal allocation ratios under matched training loss, the analysis shows that some factors mainly change where the model is trained to, while other factors keep different internal writing structures even at the same training progress.
	
	Fourth, this paper verifies the framework through controlled experiments and real neural network tasks. Controlled experiments show that the effect of learning rate largely collapses after matched-progress comparison. In contrast, \(p\), batch size, optimizer memory, data statistics, model width, activation function, and schedule leave different allocation signatures. Experiments on CIFAR-100, ResNet-18, and facial expression recognition further show that these training-time allocation differences are reflected in weight--bias gradient structure, sample loss quantiles, hard-sample migration, convergence speed, and generalization behavior.
	
	Fifth, this paper reinterprets the continuous preconditioning exponent \(p\). It is not only a scalar hyperparameter for generalization. It is also an analytical interface for EMA-dependent coordinate allocation. The results show that smaller or larger \(p\) does not imply monotonic improvement in generalization. Different tasks, data statistics, and training stages require different signal-allocation states. Therefore, \(p\) is better understood as a probe that controls the coordinate-gain structure during training. It also provides a basis for future optimization methods that control progress and allocation separately.
	
	\section{Related Work}
	\label{sec:related_work}
	
	This section reviews several lines of work related to training-time implicit bias and information allocation. We first discuss first-order optimizers and adaptive preconditioning because they provide the basic update form in modern neural network training. We then discuss partial adaptivity and continuous preconditioning exponents, implicit bias and solution geometry, stochastic training dynamics induced by learning rate and batch size, the effects of data statistics and model structure on training signals, parameter grouping and layer-wise diagnostics, and training schedules and dynamic control. Finally, we explain how these areas relate to the information-allocation view in this paper.
	
	\subsection{First-order optimizers and adaptive preconditioning}
	\label{subsec:rw_first_order_preconditioning}
	
	First-order optimizers use mini-batch gradients as the main training signal and update parameters iteratively. Let the current mini-batch loss be \(\mathcal{L}_t(\theta_t)\). The gradient is
	\begin{equation}
		g_t
		=
		\nabla_{\theta}\mathcal{L}_t(\theta_t).
		\label{eq:rw_gradient}
	\end{equation}
	Classical SGD directly updates parameters using the current gradient \cite{robbins1951stochastic}. Momentum adds historical accumulation to the gradient direction. This reduces oscillation and accelerates movement along low-curvature directions \cite{polyak1964some}. RMSProp uses historical squared gradients to build coordinate-wise scale factors. Thus, different coordinates obtain different effective update magnitudes according to their gradient statistics \cite{geoffrey2012rmsprop}. Adam further combines first-moment and second-moment estimates. It has become a widely used adaptive optimization framework in deep learning \cite{kingma2014adam}. AdamW decouples weight decay from the adaptive gradient update. This gives a clearer separation between parameter shrinkage and gradient preconditioning \cite{loshchilov2017decoupled}.
	
	In a unified form, these methods can be written as updates determined by a first-order direction, optimizer state, and a preconditioning factor:
	\begin{equation}
		\theta_{t+1}
		=
		\theta_t
		-
		\eta_t P_t \tilde g_t,
		\label{eq:rw_unified_first_order}
	\end{equation}
	where \(\eta_t\) is the learning rate, \(\tilde g_t\) is the current gradient or a momentum-smoothed first-order direction, and \(P_t\) is a preconditioning factor determined by the optimizer state. When \(P_t\) is close to the identity matrix, the update is close to SGD or Momentum. When \(P_t\) changes coordinate by coordinate according to historical gradient statistics, the update behaves like RMSProp, Adam, or AdamW.
	
	Adam-like methods usually maintain first-moment and second-moment estimates:
	\begin{equation}
		m_t
		=
		\beta_1m_{t-1}
		+
		(1-\beta_1)g_t,
		\qquad
		v_t
		=
		\beta_2v_{t-1}
		+
		(1-\beta_2)g_t^2.
		\label{eq:rw_adam_m_v}
	\end{equation}
	After bias correction, Adam uses \(\hat m_t/(\sqrt{\hat v_t}+\epsilon)\) to form the coordinate-wise update direction. The first moment controls how gradient directions are accumulated over time. The second-moment estimate controls the relative gain of different coordinates. The term \(\epsilon\) provides a numerical floor for coordinates with small second moments. These designs mean that modern optimizers do not only take a scalar step along the gradient direction. They also use internal states to change the effective update structure across coordinates, parameter groups, and training stages.
	
	\subsection{Partial adaptivity and continuous preconditioning exponents}
	\label{subsec:rw_partial_adaptivity}
	
	In Adam, the second moment enters the update denominator through a square root. More generally, a continuous exponent can control the strength of second-moment preconditioning:
	\begin{equation}
		\Delta\theta_{t,i}
		=
		-
		\eta_t
		\frac{\hat m_{t,i}}
		{(\hat v_{t,i}+\epsilon)^p},
		\label{eq:rw_p_update}
	\end{equation}
	where \(i\) denotes a parameter coordinate, and \(p\) controls how strongly the second-moment estimate enters the update. When \(p=0\), the second moment does not explicitly scale the update. The update is close to a momentum-like update. When \(p=1/2\), the update is close to Adam. Intermediate values give different strengths of diagonal preconditioning.
	
	Partial adaptive methods such as Padam use this continuous exponent to build intermediate states between SGD-like and Adam-like updates. They also analyze how different levels of adaptivity affect optimization and generalization \cite{chen2020closing}. A larger \(p\) makes the update depend more strongly on the second-moment channel. Coordinates with lower second moments obtain higher relative gain. A smaller \(p\) weakens this coordinate-wise normalization and makes the update more dominated by the first-order direction. Therefore, the continuous exponent provides a unified interface for comparing SGD-like and Adam-like methods. It also provides an adjustable parameter for studying the strength of adaptive preconditioning.
	
	From the update structure, \(p\) directly controls the coordinate gain
	\begin{equation}
		a_{t,i}(p)
		=
		(\hat v_{t,i}+\epsilon)^{-p}.
		\label{eq:rw_coordinate_gain}
	\end{equation}
	When different coordinates have different second-moment statistics, changing \(p\) changes the relative update ratio among these coordinates. This property makes the continuous preconditioning exponent an important interface for observing the internal state of adaptive optimizers.
	
	\subsection{Implicit bias, solution geometry, and generalization differences}
	\label{subsec:rw_implicit_bias_geometry}
	
	Many studies show that optimizers are not neutral numerical tools. Different optimizers can optimize the same training objective but produce different training trajectories, reachable solutions, and generalization behavior. SGD, Momentum, and Adam-like methods often show different early descent speeds, stability, and final generalization in vision, language, and other deep learning tasks \cite{wilson2017marginal,pmlr-v235-cattaneo24a,zhang2024implicit,vasudeva2026rich}.
	
	Existing studies often explain implicit bias through solution geometry and optimization trajectories. Sharpness, flatness, Hessian spectra, loss landscape visualization, and local curvature are often used to describe the region where the model finally stays. They also help explain whether different optimizers tend to reach different types of solutions \cite{zaheer2018adaptive,li2018visualizing}. These metrics provide important tools for studying the relation between optimizers and generalization. They also turn implicit bias from an empirical phenomenon into a training property that can be observed and compared.
	
	Implicit bias is also related to gradient noise, mini-batch stochasticity, and training trajectory stability. Stochastic gradients do not only provide a descent direction. They also inject noise and induce path selection during training. Learning rate, batch size, momentum, and preconditioning all change the scale, direction, and persistence of this stochastic trajectory. Thus, implicit bias can appear as a geometric preference in the final solution space. It can also appear as a way in which the model explores parameter space during training.
	
	\subsection{Learning rate, batch size, and stochastic training dynamics}
	\label{subsec:rw_lr_batch_stochastic_dynamics}
	
	The learning rate is one of the most important control variables in deep network training. It directly controls the global scale of the current update. Through multi-step iteration, it also affects the region of the loss surface reached by the model. A larger learning rate usually produces larger parameter displacement and stronger trajectory feedback. A smaller learning rate emphasizes local stable updates and late-stage convergence. Warmup, step decay, cosine annealing, and cyclic learning rates are now common components of modern training recipes.
	
	With fixed gradients and fixed optimizer states, the learning rate scales the current update as a scalar. However, in the multi-step training of nonlinear networks, the learning rate changes the next parameter position. This further changes later residuals, gradients, activation distributions, and optimizer states. Therefore, the learning rate is both a progress controller for the current step and an important source of the long-term training trajectory.
	
	Batch size affects the stochastic observation of the data distribution by mini-batches. A smaller batch size gives stronger gradient noise and higher sample-level fluctuation. A larger batch size reduces the variance of gradient estimates and makes the statistics observed in each step closer to the full data distribution. For sparse features or rare samples, batch size also changes the probability that they are observed in a mini-batch. Therefore, batch size affects not only computational efficiency. It also changes how sparse signals enter gradients, second-moment estimates, and parameter updates.
	
	Learning rate and batch size jointly determine the noise scale, update stability, and convergence rhythm of training. Together with optimizer momentum, preconditioning, and weight decay, they form the main control factors of training dynamics.
	
	\subsection{Data statistics, model capacity, and activation gating}
	\label{subsec:rw_data_capacity_activation}
	
	Data statistics determine the types of information that training signals can use. Dense stable features, sparse useful features, and spuriously correlated features enter training with different frequencies and different gradient patterns. For the same model and optimizer, different data distributions can lead to very different gradient structures, sample fitting orders, and generalization behavior. Data augmentation, class imbalance, long-tail distributions, noisy labels, and spurious correlations all reflect statistical differences in training signals.
	
	Model capacity determines how many degrees of freedom the network can use to absorb error. Wider layers, deeper structures, and richer module combinations provide more input-dependent transformation paths. These paths allow the model to fit data through more complex representations. Smaller or constrained models rely more on existing features, normalization parameters, biases, classifier heads, or other calibration paths for local correction. Therefore, model size and structure affect not only function approximation. They also affect which parameter paths can receive error signals.
	
	Activation functions and normalization layers further change gradient propagation and local gating. ReLU, GELU, LeakyReLU, and tanh have different derivative forms. They affect which samples, channels, and input regions enter the effective gradient. BatchNorm, LayerNorm, and other normalization methods change the distribution of intermediate representations through scale and shift calibration. They also make normalization parameters important learnable adjustment paths in modern networks. Thus, data statistics, model capacity, and activation gating jointly determine how training signals can be observed, propagated, and written.
	
	\subsection{Parameter groups, normalization parameters, and layer-wise diagnostics}
	\label{subsec:rw_parameter_groups_layerwise_diagnostics}
	
	Modern deep learning practice widely uses parameter groups. A typical example is group-wise weight decay. Many training recipes apply weight decay only to weights in convolutional and linear layers. They do not apply the same regularization to biases, BatchNorm \(\gamma,\beta\), LayerNorm parameters, or other one-dimensional calibration parameters. This practice is especially common with decoupled weight decay methods such as AdamW because gradient updates, parameter shrinkage, and adaptive preconditioning can be set separately.
	
	Parameter grouping reflects functional differences among parameter types. Convolution kernels, linear weights, attention projections, and FFN weights usually perform input-dependent transformations. They determine how information is combined across channels, spatial positions, or tokens. Explicit biases, normalization shifts, normalization scales, and classifier heads more often participate in output shift, scale calibration, local boundary correction, or task-specific mapping. These parameters do not have exactly the same mathematical role. However, they can form training paths that differ from the main weight matrices.
	
	Layer-wise diagnostics are also widely used to monitor deep network training. Research and engineering practice often record layer-wise gradient norms, update norms, parameter norms, update-to-weight ratios, activation statistics, and normalization statistics. These quantities are used to check exploding gradients, vanishing gradients, numerical instability, differences in layer-wise learning speed, and abnormal parameter updates. Layer-wise learning rate, LARS, LAMB, and parameter-group learning rates further show that different layers and parameter groups can have different update scales and training needs.
	
	These practices provide rich tools for observing the training process. They show that parameter update is not only a global scalar process. It can be decomposed into layers, modules, parameter types, and statistical channels for analysis.
	
	\subsection{Training schedules and dynamic optimization strategies}
	\label{subsec:rw_training_control}
	
	Besides fixed optimizer rules, schedules during training can strongly affect model behavior. Learning-rate warmup, step decay, cosine annealing, cyclic learning rate, early stopping, and optimizer switching all affect the final model by changing update scale, exploration range, and convergence rhythm across training stages. They usually do not change the loss function itself. Instead, they change the time structure of training.
	
	Learning-rate schedules control how the speed of training progress changes over time. Warmup is often used to avoid early instability. Decay is used to reduce the late-stage update size and improve convergence stability. Cyclic learning rates and cosine schedules maintain some exploration through periodic or smooth changes. Optimizer switching changes the update rule more discretely. For example, training may use an Adam-like method early for fast loss reduction and then switch to an SGD-like method later for different convergence behavior.
	
	Dynamic optimization strategies can also act on weight decay, momentum, batch size, or preconditioning states. Different scheduled variables change the order in which training signals enter the model. The model can then experience different update geometries and statistical states at different stages. Therefore, training is determined not only by final hyperparameter values. It is also determined by the path of these hyperparameters over time.
	
	\subsection{Summary and position of this paper}
	\label{subsec:rw_summary}
	
	In summary, existing studies describe implicit bias and training dynamics in neural networks from many directions. First-order optimizers and adaptive preconditioning define the basic update rules. Partial adaptive methods provide a continuous interface for controlling second-moment preconditioning strength. Studies on implicit bias and solution geometry explain why different optimizers may reach different types of solutions. Studies on learning rate, batch size, and stochastic training dynamics show that training progress, noise, and trajectory feedback can affect generalization. Data statistics, model capacity, and activation gating determine the observability and writable paths of training signals. Parameter grouping and layer-wise diagnostics provide tools for observing the training states of different parameter groups. Schedules show that the temporal order of training states is also important.
	
	Based on these works, this paper uses training-time information allocation as a common observation layer. Unlike analyses that mainly focus on final-solution geometry or the performance difference of a single optimizer, this paper focuses on how error signals are written into parameter paths, coordinate channels, and sample regions during training. The continuous preconditioning exponent \(p\), learning rate, weight decay, batch size, optimizer memory, data statistics, model structure, activation function, and schedule are all treated as possible sources that change information allocation. The following sections analyze the allocation signatures of these heterogeneous sources through diagnostics such as gradient demand, update injection, EMA-induced gain, channel update ratio, and sample-level loss distribution.
	
	
	\section{Information Allocation Methodology}
	\label{sec:information_allocation_methodology}
	
	This section gives a path-level view for analyzing optimizer implicit bias from the training process itself. Existing analyses of adaptive optimizers usually emphasize coordinate-wise preconditioning. Adam-like methods use exponential moving averages of first moments and second moments to change the effective learning rate of different coordinates. This explanation is necessary, but it stays at the coordinate level. Parameters in neural networks are not unstructured coordinates. Different parameter types correspond to different Jacobian paths. Therefore, the same error signal enters the model through different paths.
	
	This paper focuses on the smallest and most stable decomposition: weight-like parameters and bias-like parameters. Weight-like parameters usually participate in residual correction through input-dependent transformations. Bias-like parameters usually participate in residual correction through offset, scale, or calibration paths. This paper calls the relative writing of training signals between these two parameter paths weight--bias information allocation. This view does not claim that weight and bias explain all network dynamics. Instead, it uses them as a minimal observable projection of training-time implicit bias.
	
	The main line of this section is as follows. First, we define gradient-demand allocation, actual-update allocation, and optimizer-induced allocation. Then, in a minimal linear model, we show that weight and bias gradients come from different projections of the same residual. Next, we substitute a preconditioned update into residual dynamics. This shows how the optimizer changes the relative strength of input-conditioned correction and offset correction. We then give a unified training recursion. This recursion shows where learning rate, weight decay, batch size, \(\beta_1\), \(\beta_2\), \(\epsilon\), \(p\), data statistics, model width, activation function, and schedule enter weight--bias allocation. Finally, we extend the analysis to modern networks and give testable predictions.
	
	This paper provides a minimal observable projection. It is a training-time observation layer that is common in neural networks, easy to record, and connected to clear residual paths. Other decompositions, such as layer-wise, module-wise, frequency-domain, or sample-group decompositions, can reveal complementary mechanisms. We choose the weight--bias decomposition because it gives an analysis interface that is stable across model architectures and connects optimizer statistics with parameter-path dynamics.
	
	\subsection{Training-time allocation diagnostics}
	\label{subsec:training_time_allocation_diagnostics}
	
	Let the model parameters be divided into a weight-like group \(\theta_w\) and a bias-like group \(\theta_b\). At step \(t\), the mini-batch loss is \(\mathcal{L}_t(\theta_t)\), and the gradient is
	\begin{equation}
		g_t
		=
		\nabla_\theta \mathcal{L}_t(\theta_t)
		=
		(g_{w,t},g_{b,t}).
		\label{eq:chapter3_gradient_group}
	\end{equation}
	The optimizer uses the gradient and the internal state \(s_t\) to produce an update
	\begin{equation}
		\theta_{t+1}
		=
		\theta_t+\Delta\theta_t,
		\qquad
		\Delta\theta_t
		=
		-\eta_t\mathcal{P}_t(g_t;s_t),
		\label{eq:chapter3_update_operator}
	\end{equation}
	where \(\eta_t\) is the learning rate, and \(\mathcal{P}_t\) denotes the gradient transformation determined by the optimizer state. Correspondingly,
	\begin{equation}
		\Delta\theta_t
		=
		(\Delta w_t,\Delta b_t).
		\label{eq:chapter3_update_group}
	\end{equation}
	
	This paper separates three types of training-time quantities. The first is gradient-demand allocation:
	\begin{equation}
		R_g(t)
		=
		\frac{\|g_{w,t}\|_2}
		{\|g_{b,t}\|_2+\epsilon_g}.
		\label{eq:chapter3_Rg}
	\end{equation}
	It describes the current gradient demand produced jointly by the model, data, and residual. The second is actual-update allocation:
	\begin{equation}
		R_u(t)
		=
		\frac{\|\Delta w_t\|_2}
		{\|\Delta b_t\|_2+\epsilon_u}.
		\label{eq:chapter3_Ru}
	\end{equation}
	It describes the update ratio that the optimizer actually writes into the model. The third is optimizer-induced allocation:
	\begin{equation}
		R_{\mathrm{opt}}(t)
		=
		\frac{R_u(t)}
		{R_g(t)+\epsilon_r}.
		\label{eq:chapter3_Ropt}
	\end{equation}
	It describes whether the optimizer further changes the relative writing between the weight-like path and the bias-like path under a given gradient demand. In practice, the corresponding logarithmic form can also be used:
	\begin{equation}
		L_{\mathrm{opt}}(t)
		=
		\log R_u(t)-\log R_g(t).
		\label{eq:chapter3_log_Ropt}
	\end{equation}
	This form turns multiplicative deviation into additive deviation. It also avoids excessive amplification of ratio-of-ratios when the denominator is very small. If \(L_{\mathrm{opt}}(t)>0\), the actual update is more biased toward the weight-like path than the gradient demand. If \(L_{\mathrm{opt}}(t)<0\), it is more biased toward the bias-like path.
	
	These three quantities correspond to different levels. \(R_g(t)\) mainly reflects the geometric relation between the current residual and the Jacobian. \(R_u(t)\) reflects the actual update after optimizer state, preconditioning, and learning rate. \(R_{\mathrm{opt}}(t)\) or \(L_{\mathrm{opt}}(t)\) is used to observe whether the optimizer transformation changes the conversion from gradient demand to update. Gradient norms, update norms, EMA statistics, and effective gains in later experiments can all be interpreted in this hierarchy.
	
	\subsection{Minimal path decomposition: weight and bias}
	\label{subsec:minimal_path_decomposition}
	
	To show that weight and bias are not arbitrary groups, we first consider a single-channel linear model:
	\begin{equation}
		z
		=
		Xw+b\mathbf{1},
		\label{eq:chapter3_linear_model}
	\end{equation}
	where \(X\in\mathbb{R}^{n\times d}\) is the input matrix, \(w\in\mathbb{R}^d\) is the weight, \(b\in\mathbb{R}\) is the bias, and \(\mathbf{1}\in\mathbb{R}^n\) is an all-one vector. Let the target be \(y\). The residual is
	\begin{equation}
		r
		=
		z-y
		=
		Xw+b\mathbf{1}-y.
		\label{eq:chapter3_residual}
	\end{equation}
	We use the squared loss
	\begin{equation}
		\mathcal{L}
		=
		\frac{1}{2n}\|r\|_2^2.
		\label{eq:chapter3_square_loss}
	\end{equation}
	Then the gradients of weight and bias are
	\begin{equation}
		g_w
		=
		\nabla_w\mathcal{L}
		=
		\frac{1}{n}X^\top r,
		\qquad
		g_b
		=
		\nabla_b\mathcal{L}
		=
		\frac{1}{n}\mathbf{1}^\top r.
		\label{eq:chapter3_wb_gradients}
	\end{equation}
	
	Equation \eqref{eq:chapter3_wb_gradients} gives the minimal source of weight--bias allocation. \(X^\top r\) is the projection of the residual onto input-dependent directions. \(\mathbf{1}^\top r\) is the projection of the residual onto the constant direction. It is the residual mean. Thus, the weight-like path and the bias-like path correspond to two different residual absorption paths:
	\begin{equation}
		\text{weight-like path: } X^\top r,
		\qquad
		\text{bias-like path: } \mathbf{1}^\top r.
		\label{eq:chapter3_two_paths}
	\end{equation}
	In this model, the weight-like path absorbs the input-conditioned residual component. The bias-like path absorbs the input-independent offset component. In deep networks, \(X\) and \(\mathbf{1}\) are replaced by more general Jacobians \(J_w\) and \(J_b\). However, the path difference remains.
	
	\subsection{Preconditioned residual correction}
	\label{subsec:preconditioned_residual_correction}
	
	We next consider one preconditioned update:
	\begin{equation}
		\Delta w
		=
		-\eta Q_w g_w,
		\qquad
		\Delta b
		=
		-\eta q_b g_b,
		\label{eq:chapter3_preconditioned_wb_update}
	\end{equation}
	where \(Q_w\) is the diagonal preconditioner for weight coordinates, and \(q_b\) is the scalar preconditioning factor for the bias. The residual after the update is
	\begin{align}
		r^+
		&=
		X(w+\Delta w)+(b+\Delta b)\mathbf{1}-y \notag\\
		&=
		r+X\Delta w+\Delta b\,\mathbf{1}.
		\label{eq:chapter3_residual_after_update_a}
	\end{align}
	Substituting Eq. \eqref{eq:chapter3_wb_gradients} and Eq. \eqref{eq:chapter3_preconditioned_wb_update} gives
	\begin{equation}
		r^+
		=
		r
		-
		\frac{\eta}{n}XQ_wX^\top r
		-
		\frac{\eta}{n}q_b\mathbf{1}\mathbf{1}^\top r.
		\label{eq:chapter3_residual_after_update}
	\end{equation}
	
	Define two residual correction operators:
	\begin{equation}
		K_w(Q_w)
		=
		XQ_wX^\top,
		\qquad
		K_b(q_b)
		=
		q_b\mathbf{1}\mathbf{1}^\top.
		\label{eq:chapter3_correction_operators}
	\end{equation}
	Then Eq. \eqref{eq:chapter3_residual_after_update} can be written as
	\begin{equation}
		r^+
		=
		r
		-
		\frac{\eta}{n}K_w(Q_w)r
		-
		\frac{\eta}{n}K_b(q_b)r.
		\label{eq:chapter3_operator_residual_update}
	\end{equation}
	This shows that the optimizer does not only change the step size in parameter space. It also changes the relative strength by which the same residual is absorbed by input-conditioned correction and offset correction. Accordingly, we can define a path correction ratio:
	\begin{equation}
		A_{w/b}
		=
		\frac{\|K_w(Q_w)r\|_2}
		{\|K_b(q_b)r\|_2+\epsilon_a}.
		\label{eq:chapter3_path_correction_ratio}
	\end{equation}
	In deep networks, directly computing \(K_w r\) and \(K_b r\) is usually expensive. Therefore, this paper uses \(R_g\), \(R_u\), \(R_{\mathrm{opt}}\), and EMA gain ratios as recordable path-level diagnostics.
	
	\subsection{Why matched progress is necessary but insufficient}
	\label{subsec:matched_progress_theory}
	
	The largest confounder in training-time analysis is training progress. If different training settings are compared only at a fixed epoch, then loss, accuracy, gradient norm, and the loss tail may mainly reflect how far the model has been trained. They do not necessarily reflect the internal mechanism. Therefore, matched-progress comparison is necessary.
	
	In the minimal model, the training loss only constrains the residual norm:
	\begin{equation}
		\mathcal{L}
		=
		\frac{1}{2n}\|r\|_2^2.
		\label{eq:chapter3_loss_residual_norm}
	\end{equation}
	Two models with similar training loss only have similar \(\|r\|_2\). This does not mean that the following quantities are the same:
	\begin{equation}
		X^\top r,
		\quad
		\mathbf{1}^\top r,
		\quad
		Q_w,
		\quad
		q_b,
		\quad
		m_t,
		\quad
		v_t.
		\label{eq:chapter3_unmatched_internal_quantities}
	\end{equation}
	
	More specifically, let \(\mathcal{S}_X=\operatorname{span}(X)\), \(\mathcal{S}_1=\operatorname{span}(\mathbf{1})\), and
	\begin{equation}
		\mathcal{S}
		=
		\mathcal{S}_X+\mathcal{S}_1 .
		\label{eq:chapter3_sum_subspace}
	\end{equation}
	Let \(P_{\mathcal{S}}\) be the orthogonal projection onto \(\mathcal{S}\). The residual can be safely written as
	\begin{equation}
		r
		=
		P_{\mathcal{S}}r+r_{\perp},
		\qquad
		r_{\perp}\perp \mathcal{S}.
		\label{eq:chapter3_safe_residual_decomposition}
	\end{equation}
	Matched training loss constrains \(\|r\|_2\). It does not constrain the alignment of \(P_{\mathcal{S}}r\) with the directions in \(\mathcal{S}_X\) and \(\mathcal{S}_1\). Since \(\mathcal{S}_X\) and \(\mathcal{S}_1\) are not necessarily orthogonal, this paper does not assume that \(\|r\|_2^2\) has a unique decomposition into simple squared input-conditioned energy and offset energy. Instead, we use a weaker but sufficient fact:
	\begin{equation}
		\|r\|_2 \text{ fixed}
		\nRightarrow
		\left(\|X^\top r\|_2, |\mathbf{1}^\top r|\right) \text{ fixed}.
		\label{eq:chapter3_norm_not_orientation}
	\end{equation}
	In other words, the same residual norm does not imply the same residual orientation. The same training progress does not imply the same allocation geometry.
	
	Therefore, matched-progress comparison does not remove all differences. Its role is to separate external progress differences from internal path differences. If a factor mainly acts through training progress, then both external loss distributions and internal ratios should become close after matched training loss. If a factor changes residual orientation, optimizer state, or coordinate-wise gain, then the external loss distribution may collapse, but \(R_g\), \(R_u\), \(R_{\mathrm{opt}}\), or gain ratios may still remain different.
	
	\subsection{A unified recursion for training-time allocation sources}
	\label{subsec:unified_training_recursion}
	
	A general neural network can be written as \(f_\theta(x)\). Let the output error signal of the \(i\)-th sample be \(e_i(\theta_t)\), and let the corresponding Jacobian be \(J_i(\theta_t)\). The mini-batch gradient can be written as
	\begin{equation}
		g_{B_t}(\theta_t)
		=
		\frac{1}{B}
		\sum_{i\in\mathcal{B}_t}
		J_i(\theta_t)^\top e_i(\theta_t).
		\label{eq:chapter3_minibatch_gradient_jacobian}
	\end{equation}
	An Adam-like optimizer maintains first-moment and second-moment estimates:
	\begin{equation}
		m_t
		=
		\beta_1m_{t-1}
		+
		(1-\beta_1)g_{B_t},
		\qquad
		v_t
		=
		\beta_2v_{t-1}
		+
		(1-\beta_2)g_{B_t}^2.
		\label{eq:chapter3_adam_state}
	\end{equation}
	Ignoring bias-correction notation, a coordinate-wise update with continuous preconditioning exponent \(p\) and decoupled weight decay can be written as
	\begin{equation}
		\Delta\theta_{t,i}
		=
		-\eta_t
		\frac{m_{t,i}}
		{(v_{t,i}+\epsilon)^p}
		-
		\eta_t\lambda_i\theta_{t,i}.
		\label{eq:chapter3_unified_update}
	\end{equation}
	Here \(\lambda_i\) may depend on the parameter type. For example, weight-like parameters can use nonzero weight decay, while biases or normalization affine parameters can use \(\lambda_i=0\).
	
	Equations \eqref{eq:chapter3_minibatch_gradient_jacobian}--\eqref{eq:chapter3_unified_update} give a unified entry point for analyzing training-time allocation sources. The term source is used in a broad sense. Some sources belong to the optimizer rule. Some belong to the stochastic training protocol. Some come from the data distribution or architecture. They are placed in the same framework because they all enter the same training recursion, and they can all be observed through allocation diagnostics. Different sources enter the training recursion at different positions:
	\begin{equation}
		\begin{array}{ll}
			\text{data statistics} & \rightarrow \quad (x_i,y_i),e_i,J_i,\\
			\text{model width} & \rightarrow \quad \operatorname{rank}(J_w)\text{ and weight-like paths},\\
			\text{activation function} & \rightarrow \quad J_i(\theta_t)\text{ through nonlinear gating},\\
			\text{batch size} & \rightarrow \quad g_{B_t}\text{ as stochastic gradient estimation},\\
			\beta_1 & \rightarrow \quad m_t\text{ as first-moment memory},\\
			\beta_2 & \rightarrow \quad v_t\text{ as second-moment memory},\\
			p & \rightarrow \quad (v_t+\epsilon)^{-p}\text{ as coordinate-wise gain},\\
			\epsilon & \rightarrow \quad \text{low-}v\text{ gain floor},\\
			\eta_t & \rightarrow \quad \text{global progress scale and trajectory feedback},\\
			\lambda_i & \rightarrow \quad \text{decoupled shrinkage},\\
			\text{schedule} & \rightarrow \quad \text{time dependence of }\eta_t,p_t,\lambda_t,\beta_t.
		\end{array}
		\label{eq:chapter3_sources_positions}
	\end{equation}
	Therefore, training-time allocation sources should not be treated as unrelated empirical factors. They all change gradient observation, momentum memory, second-moment memory, coordinate-wise gain, shrinkage drift, or trajectory feedback through Eq. \eqref{eq:chapter3_unified_update}. They are then projected into weight--bias allocation.
	
	\subsection{Source-wise mechanisms}
	\label{subsec:source_wise_mechanisms}
	
	This section explains how the main training-time allocation sources enter the unified recursion. Each source acts at a different position. Therefore, it should have a different matched-progress signature.
	
	\subsubsection{\(p\) as a coordinate-gain source}
	\label{subsubsec:p_coordinate_gain_source}
	
	Define the coordinate-wise gain:
	\begin{equation}
		a_{t,i}(p)
		=
		(v_{t,i}+\epsilon)^{-p}.
		\label{eq:chapter3_coordinate_gain}
	\end{equation}
	Then the update can be written as
	\begin{equation}
		\Delta\theta_{t,i}
		=
		-\eta_t a_{t,i}(p)m_{t,i}
		-
		\eta_t\lambda_i\theta_{t,i}.
		\label{eq:chapter3_update_gain_form}
	\end{equation}
	Ignoring weight decay, for two coordinates \(i\) and \(j\), we have
	\begin{equation}
		\frac{|\Delta\theta_{t,i}|}
		{|\Delta\theta_{t,j}|}
		=
		\frac{|m_{t,i}|}
		{|m_{t,j}|}
		\left(
		\frac{v_{t,j}+\epsilon}
		{v_{t,i}+\epsilon}
		\right)^p.
		\label{eq:chapter3_coordinate_update_ratio}
	\end{equation}
	Thus, \(p\) is not a global scale. It converts second-moment statistics into coordinate-wise gain geometry. Projected onto weight-like and bias-like parameter groups,
	\begin{equation}
		R_u(p)
		=
		\frac{\|a_w(p)\odot m_w\|_2}
		{\|a_b(p)\odot m_b\|_2+\epsilon_u}.
		\label{eq:chapter3_Ru_p}
	\end{equation}
	If the external loss distribution becomes close after matched training loss, but the gain ratio or \(R_{\mathrm{opt}}\) remains different, then \(p\) changes the internal allocation geometry used to reach a similar empirical risk.
	
	\subsubsection{Sparse coordinates and low second moments}
	\label{subsubsec:sparse_coordinates_low_v}
	
	Consider a weight coordinate \(j\). Its mini-batch gradient is
	\begin{equation}
		g_{B,j}
		=
		\frac{1}{B}
		\sum_{i\in\mathcal{B}}
		x_{ij}e_i.
		\label{eq:chapter3_sparse_gradient}
	\end{equation}
	If \(x_{ij}\) is nonzero with probability \(\pi_j\), and if the squared magnitude of \(x_{ij}e_i\) when nonzero is \(\sigma_j^2\), then under ignored sign correlation and locally stationary statistics, the second moment in a stable stage can be approximated as
	\begin{equation}
		\mathbb{E}[v_j]
		\approx
		\pi_j\sigma_j^2.
		\label{eq:chapter3_sparse_v_expectation}
	\end{equation}
	The corresponding gain is
	\begin{equation}
		a_j(p)
		=
		(v_j+\epsilon)^{-p}
		\approx
		(\pi_j\sigma_j^2+\epsilon)^{-p}.
		\label{eq:chapter3_sparse_gain}
	\end{equation}
	When \(\epsilon\) is not dominant,
	\begin{equation}
		a_j(p)
		\propto
		\pi_j^{-p}.
		\label{eq:chapter3_sparse_gain_pi}
	\end{equation}
	
	This approximation shows that input sparsity can enter coordinate-wise gain through the second-moment estimate. A larger \(p\) amplifies the relative gain of low-\(v\) coordinates. However, low \(v\) does not mean useful signal. It can correspond to a sparse useful feature. It can also correspond to sparse noise or a spurious feature. Therefore, the effect of \(p\) should be understood as changing the relative gain of low-second-moment coordinates. It should not be interpreted as direct selection of the correct feature.
	
	\subsubsection{Batch size as a stochastic-observation source}
	\label{subsubsec:batch_stochastic_observation}
	
	Batch size changes stochastic gradient estimation. For a sparse feature with occurrence probability \(\pi_j\), under an independent-sampling approximation, the probability that this feature is completely absent in one batch is
	\begin{equation}
		P_{\mathrm{absent}}(B)
		=
		(1-\pi_j)^B.
		\label{eq:chapter3_batch_absent_probability}
	\end{equation}
	The probability of observing it at least once is
	\begin{equation}
		P_{\mathrm{seen}}(B)
		=
		1-(1-\pi_j)^B.
		\label{eq:chapter3_batch_seen_probability}
	\end{equation}
	Therefore, a large batch mainly reduces mini-batch observation sparsity. It does not change the sparsity of the dataset itself.
	
	On the other hand, let the single-sample variable be \(h_j=x_je\). Ignoring finite-population correction, the variance of the mini-batch gradient is approximately
	\begin{equation}
		\operatorname{Var}(g_{B,j})
		\approx
		\frac{1}{B}\operatorname{Var}(h_j).
		\label{eq:chapter3_batch_variance}
	\end{equation}
	Thus, batch size changes both observation sparsity and gradient noise. A small batch is more likely to produce long absence and occasional spikes for rare coordinates. A large batch observes sparse features more continuously and reduces gradient noise. These two effects can influence \(v_j\) and gain in different directions. Therefore, the allocation pattern induced by batch size does not have to be monotonic. This is also the key difference between batch size and learning rate. Learning rate mainly changes step scale, while batch size changes the gradient statistics observed by the optimizer at each step.
	
	\subsubsection{\(\beta_2\) as second-moment memory}
	\label{subsubsec:beta2_memory_source}
	
	The second moment can be expanded as
	\begin{equation}
		v_t
		=
		(1-\beta_2)
		\sum_{s=1}^{t}
		\beta_2^{t-s}
		g_{B_s}^2.
		\label{eq:chapter3_beta2_expansion}
	\end{equation}
	Its effective memory length in steps is approximately
	\begin{equation}
		\tau_2
		\approx
		\frac{1}{1-\beta_2}.
		\label{eq:chapter3_beta2_time_constant}
	\end{equation}
	If each step uses batch size \(B\), then the same EMA window covers about
	\begin{equation}
		B\tau_2
		\approx
		\frac{B}{1-\beta_2}
		\label{eq:chapter3_beta2_sample_time_constant}
	\end{equation}
	samples in terms of processed examples. Thus, changing batch size also changes the sample-scale meaning of the same \(\beta_2\).
	
	Therefore, \(\beta_2\) decides which historical squared gradients enter the current second-moment estimate. The exponent \(p\) decides how the current second-moment estimate becomes a gain:
	\begin{equation}
		a_i(p,\beta_2)
		=
		\left[
		(1-\beta_2)
		\sum_{s=1}^{t}
		\beta_2^{t-s}
		g_{B_s,i}^2
		+
		\epsilon
		\right]^{-p}.
		\label{eq:chapter3_beta2_p_gain}
	\end{equation}
	Therefore, \(\beta_2\) is a second-moment memory source, and \(p\) is a coordinate-gain exponent. Together, they determine the EMA-channel gain geometry.
	
	\subsubsection{\(\beta_1\) as first-moment memory}
	\label{subsubsec:beta1_memory_source}
	
	The first moment can be expanded as
	\begin{equation}
		m_t
		=
		(1-\beta_1)
		\sum_{s=1}^{t}
		\beta_1^{t-s}
		g_{B_s}.
		\label{eq:chapter3_beta1_expansion}
	\end{equation}
	Thus, \(\beta_1\) changes how gradient demand accumulates into a persistent update direction. For parameter groups with stable directions, a larger \(\beta_1\) strengthens the accumulation of historical directions. For parameter groups with fluctuating directions, momentum may cause cancellation. Projected onto weight--bias allocation, \(\beta_1\) mainly affects
	\begin{equation}
		R_u
		=
		\frac{\|a_w\odot m_w\|_2}
		{\|a_b\odot m_b\|_2+\epsilon_u},
		\label{eq:chapter3_beta1_Ru}
	\end{equation}
	and does not directly change the gain \(a=(v+\epsilon)^{-p}\). Therefore, under matched progress, \(\beta_1\) can show similar external metrics while keeping differences in \(R_{\mathrm{opt}}\) or update ratios.
	
	\subsubsection{\(\epsilon\) as a gain-floor source}
	\label{subsubsec:epsilon_gain_floor}
	
	For two coordinates \(i,j\), the gain ratio is
	\begin{equation}
		\frac{a_i}{a_j}
		=
		\left(
		\frac{v_j+\epsilon}
		{v_i+\epsilon}
		\right)^p.
		\label{eq:chapter3_epsilon_gain_ratio}
	\end{equation}
	When \(v_i,v_j\gg \epsilon\),
	\begin{equation}
		\frac{a_i}{a_j}
		\approx
		\left(
		\frac{v_j}{v_i}
		\right)^p.
		\label{eq:chapter3_epsilon_small}
	\end{equation}
	When \(v_i,v_j\ll \epsilon\),
	\begin{equation}
		\frac{a_i}{a_j}
		\approx
		1.
		\label{eq:chapter3_epsilon_large}
	\end{equation}
	Therefore, \(\epsilon\) is a low-\(v\) gain floor. It limits the amplification of low-second-moment coordinates by \(p\). If \(\epsilon\) is large enough, coordinate-wise preconditioning is flattened, and the allocation effect of \(p\) is weakened.
	
	\subsubsection{Learning rate and weight decay}
	\label{subsubsec:lr_wd_sources}
	
	The learning rate is mainly a global progress scale in the current step. If the preconditioner \(P_t\) is fixed, then
	\begin{equation}
		\frac{\|\Delta w_t\|_2}
		{\|\Delta b_t\|_2}
		=
		\frac{\eta_t\|P_{w,t}g_{w,t}\|_2}
		{\eta_t\|P_{b,t}g_{b,t}\|_2}
		=
		\frac{\|P_{w,t}g_{w,t}\|_2}
		{\|P_{b,t}g_{b,t}\|_2}.
		\label{eq:chapter3_lr_ratio_cancel}
	\end{equation}
	Thus, the learning rate does not directly change the current-step relative preconditioning gain. Its implicit effect mainly comes from multi-step trajectory feedback:
	\begin{equation}
		\theta_{t+1}
		=
		\theta_t
		-
		\eta_tP_tg_t,
		\label{eq:chapter3_lr_trajectory}
	\end{equation}
	which then changes \(r_{t+1}\), \(J_{t+1}\), \(g_{t+1}\), \(m_{t+1}\), and \(v_{t+1}\). This explains why the learning rate can strongly change external metrics at a fixed epoch, but tends to collapse more easily after matched-progress comparison.
	
	For decoupled weight decay, the update of weight-like parameters can be written as
	\begin{equation}
		\Delta w
		=
		-\eta Q_w g_w
		-
		\eta\lambda w,
		\qquad
		\Delta b
		=
		-\eta q_b g_b.
		\label{eq:chapter3_weight_decay_update}
	\end{equation}
	Substituting this into the minimal model gives
	\begin{equation}
		r^+
		=
		r
		-
		\frac{\eta}{n}XQ_wX^\top r
		-
		\frac{\eta}{n}q_b\mathbf{1}\mathbf{1}^\top r
		-
		\eta\lambda Xw.
		\label{eq:chapter3_weight_decay_residual}
	\end{equation}
	The last term is shrinkage drift. It is not a correction directly generated by the current residual. Therefore, weight decay is a parameter-shrinkage source. It changes the sustainable expression magnitude of the weight-like path and indirectly affects later residuals and gradient demand. Note that Eq. \eqref{eq:chapter3_weight_decay_update} corresponds to decoupled weight decay. If coupled \(L_2\) regularization is used, \(\lambda w\) enters the gradient first and is then affected by the preconditioner. This paper uses the decoupled form to separate residual-driven correction from parameter shrinkage.
	
	\subsubsection{Data statistics, model width, and activation function}
	\label{subsubsec:data_width_activation_sources}
	
	Data statistics determine what information can be captured by gradient observation. For input coordinate \(j\), the key quantities are
	\begin{equation}
		\pi_j
		=
		P(x_j\neq0),
		\qquad
		\mu_j
		=
		\mathbb{E}[x_je],
		\qquad
		\sigma_j^2
		=
		\mathbb{E}[(x_je)^2].
		\label{eq:chapter3_data_statistics}
	\end{equation}
	Here \(\pi_j\) controls observation frequency, \(\mu_j\) controls whether the channel provides a systematic gradient direction, and \(\sigma_j^2\) controls the strength of this channel in the second-moment estimate. The difference among dense useful, sparse useful, and sparse spurious channels comes from differences in these statistics. Data structure does not only change optimization difficulty. It also changes the information that can be allocated.
	
	Model width changes the number and effective dimension of weight-like paths. For a general network,
	\begin{equation}
		g_w
		=
		J_w^\top e,
		\qquad
		g_b
		=
		J_b^\top e.
		\label{eq:chapter3_general_jacobian_wb}
	\end{equation}
	Larger width usually increases the column space or effective rank of \(J_w\). It allows the residual to project onto more input-conditioned directions. This is not a pointwise guarantee for every parameter state. It means that width changes the available input-conditioned parameter paths and can therefore change the projection geometry of \(J_w^\top e\). Thus, width mainly changes gradient-demand allocation:
	\begin{equation}
		R_g
		=
		\frac{\|J_w^\top e\|_2}
		{\|J_b^\top e\|_2+\epsilon_g}.
		\label{eq:chapter3_width_Rg}
	\end{equation}
	Therefore, width is a model-update-freedom source. It is not only a scalar optimization parameter.
	
	Activation functions change nonlinear gating. For one hidden unit,
	\begin{equation}
		h=\phi(u),
		\qquad
		u=w^\top x+b.
		\label{eq:chapter3_activation_unit}
	\end{equation}
	If the backpropagated error signal is \(\delta\), then
	\begin{equation}
		g_w
		=
		\delta\phi'(u)x,
		\qquad
		g_b
		=
		\delta\phi'(u).
		\label{eq:chapter3_activation_gradients}
	\end{equation}
	ReLU uses \(\phi'(u)=\mathbf{1}_{u>0}\), while tanh uses \(\phi'(u)=1-\tanh^2(u)\). Different activation functions change which samples and features enter the Jacobian through \(\phi'(u)\). Thus, they change the residual projection geometry. The activation function is an architecture-induced gating source.
	
	\subsubsection{Schedule as trajectory ordering}
	\label{subsubsec:schedule_trajectory_ordering}
	
	If \(p\), the learning rate, or weight decay changes over time, then the update is
	\begin{equation}
		\Delta\theta_{t,i}
		=
		-\eta_t
		\frac{m_{t,i}}
		{(v_{t,i}+\epsilon)^{p_t}}
		-
		\eta_t\lambda_{t,i}\theta_{t,i}.
		\label{eq:chapter3_schedule_update}
	\end{equation}
	A schedule does not only change final hyperparameter values. It changes the temporal order of different allocation states. A larger \(p_t\) strengthens coordinate-wise selectivity. A smaller \(p_t\) weakens second-moment-induced gain differences. A learning-rate schedule changes global progress speed. A weight-decay schedule changes shrinkage pressure. Together, these schedules determine which paths absorb residuals in early and late training. Therefore, a schedule is trajectory-mediated allocation control. It is not a static factor independent of the training state.
	
	\subsection{Extension to modern networks}
	\label{subsec:modern_network_extension}
	
	In deep networks, \(X\) and \(\mathbf{1}\) in the minimal linear model correspond to more general Jacobian paths. Given an output error signal \(e\), the gradients of weight-like and bias-like parameter groups can be written as
	\begin{equation}
		g_w
		=
		J_w^\top e,
		\qquad
		g_b
		=
		J_b^\top e.
		\label{eq:chapter3_deep_wb_jacobian}
	\end{equation}
	Here \(J_w\) is the Jacobian of weight-like parameters, and \(J_b\) is the Jacobian of bias-like parameters. Similar to the minimal model, \(J_w^\top e\) describes the projection of the error signal onto input-conditioned parameter paths. \(J_b^\top e\) describes the projection of the error signal onto offset, scale, or calibration paths.
	
	This paper uses the following approximate grouping. Weight-like parameters include weights in Linear, Convolution, attention projection, and FFN/MLP blocks. Bias-like parameters include explicit biases and normalization affine parameters such as those in BatchNorm and LayerNorm. The normalization scale parameter is not the same as an additive bias. It is placed in the bias-like group because it provides a local scale or shift calibration path, and because it is often excluded from weight decay together with biases in practice. The goal of this paper is not to prove that all bias-like parameters are mathematically equivalent. The goal is to build a path-level observable that is stable across network architectures.
	
	\begin{table}[t]
		\centering
		\caption{Weight-like and bias-like parameter groups used in this work.}
		\label{tab:chapter3_modern_parameter_groups}
		\begin{tabular}{llll}
			\toprule
			Module & Parameter & Main role & Group \\
			\midrule
			Linear / Fully-connected & weight & input-conditioned linear map & weight-like \\
			Linear / Fully-connected & bias & output offset & bias-like \\
			Convolution & weight & spatial/channel transformation & weight-like \\
			Convolution & bias & channel offset & bias-like \\
			BatchNorm & scale \(\gamma\) & scale calibration & bias-like \\
			BatchNorm & shift \(\beta\) & shift calibration & bias-like \\
			LayerNorm & scale \(\gamma\) & feature-scale calibration & bias-like \\
			LayerNorm & shift \(\beta\) & feature-shift calibration & bias-like \\
			Attention projection & weight & token/channel projection & weight-like \\
			Attention projection & bias & projection offset & bias-like \\
			FFN / MLP block & weight & nonlinear feature transformation & weight-like \\
			FFN / MLP block & bias & hidden/output offset & bias-like \\
			\bottomrule
		\end{tabular}
	\end{table}
	
	Based on this grouping, experiments can record \(R_g\), \(R_u\), \(R_{\mathrm{opt}}\), second-moment ratios, gain ratios, and channel-level ratios. These statistics are not new optimizer states. They are standard training statistics reorganized under the weight--bias path decomposition.
	
	\subsection{Testable predictions}
	\label{subsec:chapter3_testable_predictions}
	
	The above analysis gives the following testable predictions.
	
	First, matched-progress comparison should strongly reduce differences in external metrics. However, it does not have to remove differences in allocation ratios. The reason is that training loss mainly constrains residual norm. It does not constrain residual orientation, Jacobian projection, optimizer memory, or coordinate-wise gain.
	
	Second, the effect of \(p\) should mainly remain in gain ratios, update ratios, and \(R_{\mathrm{opt}}\). Because \(p\) directly acts on \((v+\epsilon)^{-p}\), it changes the mapping from second-moment statistics to coordinate-wise gain.
	
	Third, learning-rate differences should collapse more easily after matched training loss. The learning rate is mainly a global scale in the current step. Its main effect comes from trajectory feedback and training progress.
	
	Fourth, batch size should produce non-monotonic or even crossing allocation patterns. This is because batch size changes both \(P_{\mathrm{absent}}(B)=(1-\pi)^B\) and \(\operatorname{Var}(g_B)\propto 1/B\). The two effects can act in opposite directions on sparse channels.
	
	Fifth, \(\beta_1\) and \(\beta_2\) should change first-moment memory and second-moment memory, respectively. The former mainly affects the conversion from gradient demand to persistent update direction. The latter mainly affects how squared-gradient history enters the gain geometry.
	
	Sixth, \(\epsilon\) should flatten gain differences among low-\(v\) coordinates. When \(v\ll\epsilon\), the gain ratio is close to \(1\), and the coordinate-selective effect of \(p\) is weakened.
	
	Seventh, data statistics, model width, and activation function are more likely to change \(R_g\). They change the structure of \(J^\top e\), that is, gradient demand itself. In contrast, \(p\), \(\beta_2\), and \(\epsilon\) more directly change the conversion from optimizer state to update.
	
	Eighth, schedules should change the temporal order of allocation. They should not be viewed only as final-state changes. Dynamic \(p_t\), \(\eta_t\), or \(\lambda_t\) changes which parameter paths absorb the residual in early and late training.
	
	In summary, weight--bias allocation provides a minimal training-time observation layer. Different training-time allocation sources do not have to produce the same external behavior. Their difference lies in where they enter the training recursion. They finally appear as differences in progress, gradient demand, optimizer memory, coordinate-wise gain, shrinkage, Jacobian paths, or trajectory ordering. The next section verifies these predictions using matched-progress diagnostics and real-task statistics.
	
	\section{Experiments and Discussion}
	
	\subsection{A source map of training-time information allocation}
	\label{subsec:wilson_source_map}
	
	The previous section defined weight--bias allocation as a minimal observable projection of training-time implicit bias. This section uses a controlled synthetic task to test whether this projection can separate the effects of different training factors. The goal is not to find one optimal hyperparameter. It is also not to prove that one setting is better for all tasks. Instead, we ask a mechanistic question. After training progress is controlled, do different training factors still leave separable information-allocation signatures? If the effect of a factor mostly disappears after matched-progress comparison, then it is closer to a progress-dominant source. If external loss is already similar but internal ratios remain separated, then the factor is more likely to change the internal allocation geometry of the training signal.
	
	The task contains three types of input statistical channels: dense informative channels, sparse informative channels, and sparse noisy or spurious channels. This design allows us to observe not only global loss and accuracy, but also how the optimizer writes training signals into different statistical channels. Each run records two types of variables. The first type includes external training-progress variables, such as training loss, test accuracy, test-loss mean, and sample loss quantiles. The second type includes internal allocation variables, such as the weight--bias gradient ratio \(R_g\), the update-to-gradient conversion \(R_u/R_g\), the bias/weight gain ratio, and the gain ratio and update ratio of noise to dense channels. The former describes where the model has moved. The latter describes through which internal paths the model has moved.
	
	To avoid mixing mechanism effects with training progress in fixed-epoch comparison, this paper reports both fixed-epoch comparison and matched-progress comparison. The fixed-epoch view keeps the overall result under a real training recipe. It is useful for showing the combined effect of a training factor on the final state. The matched-progress view compares different runs at similar training loss. It is useful for judging whether an effect is only a difference in training progress. This paper combines these two views. We first use fixed-epoch results to show that external metrics are easily confounded by training progress. We then use matched-progress results to identify which internal allocation ratios remain.
	
\begin{figure}[t]
		\centering
		\begin{subfigure}[t]{0.48\linewidth}
			\centering
			\begin{tikzpicture}
				\begin{axis}[
					width=\linewidth,
					height=0.70\linewidth,
					xlabel={$p$},
					ylabel={Train loss},
					grid=both,
					tick label style={font=\scriptsize},
					label style={font=\small}
					]
					\addplot+[mark=*]
					table[x=p,y=TrainLoss,col sep=comma]
					{wilsonfig1fixedepochpsweep.csv};
				\end{axis}
			\end{tikzpicture}
			\caption{Effect of $p$ on training progress at a fixed epoch.}
		\end{subfigure}
		\hfill
		\begin{subfigure}[t]{0.48\linewidth}
			\centering
			\begin{tikzpicture}
				\begin{axis}[
					width=\linewidth,
					height=0.70\linewidth,
					xlabel={$p$},
					ylabel={Test loss $p95$},
					grid=both,
					tick label style={font=\scriptsize},
					label style={font=\small}
					]
					\addplot+[mark=*]
					table[x=p,y=TestLossP95,col sep=comma]
					{wilsonfig1fixedepochpsweep.csv};
				\end{axis}
			\end{tikzpicture}
			\caption{Effect of $p$ on the test tail at a fixed epoch.}
		\end{subfigure}
		
		\vspace{0.3em}
		
		\begin{subfigure}[t]{0.48\linewidth}
			\centering
			\begin{tikzpicture}
				\begin{axis}[
					width=\linewidth,
					height=0.70\linewidth,
					xlabel={Learning rate},
					ylabel={Train loss},
					xmode=log,
					grid=both,
					tick label style={font=\scriptsize},
					label style={font=\small}
					]
					\addplot+[mark=*]
					table[x=lr,y=TrainLoss,col sep=comma]
					{wilsonfig1fixedepochlrsweep.csv};
				\end{axis}
			\end{tikzpicture}
			\caption{Effect of learning rate on training progress at a fixed epoch.}
		\end{subfigure}
		\hfill
		\begin{subfigure}[t]{0.48\linewidth}
			\centering
			\begin{tikzpicture}
				\begin{axis}[
					width=\linewidth,
					height=0.70\linewidth,
					xlabel={Learning rate},
					ylabel={Test loss $p95$},
					xmode=log,
					grid=both,
					tick label style={font=\scriptsize},
					label style={font=\small}
					]
					\addplot+[mark=*]
					table[x=lr,y=TestLossP95,col sep=comma]
					{wilsonfig1fixedepochlrsweep.csv};
				\end{axis}
			\end{tikzpicture}
			\caption{Effect of learning rate on the test tail at a fixed epoch.}
		\end{subfigure}
		\caption{Progress confounding under fixed-epoch comparison. Changing $p$ and changing the learning rate both strongly change training progress. Therefore, final-epoch loss, accuracy, or tail loss alone cannot separate progress effects from allocation effects.}
		\label{fig:wilson_fixed_epoch}
	\end{figure}
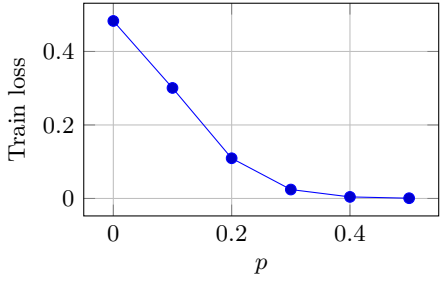
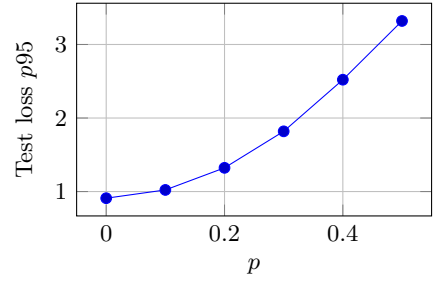
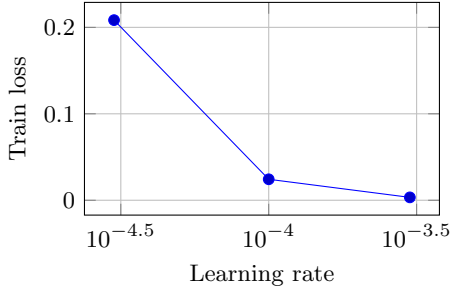
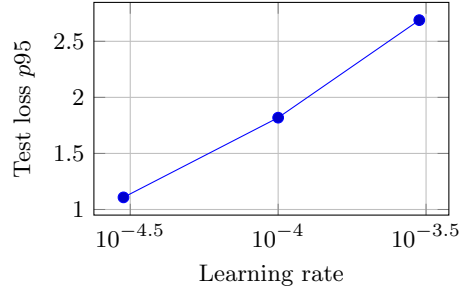

	\paragraph{Fixed-epoch comparison shows that training progress is the main confounder.}
	We first consider the result of changing \(p\) with fixed learning rate and weight decay. As shown in Fig.~\ref{fig:wilson_fixed_epoch}, increasing \(p\) from \(0\) to \(0.5\) at the final epoch strongly reduces training loss, while the test-loss tail increases. For example, when \(p=0\), the training loss is about \(0.483\), and the test-loss \(p95\) is about \(0.910\). When \(p=0.5\), the training loss decreases to about \(0.001\), while the test-loss \(p95\) increases to about \(3.320\). This means that the larger test tail observed at a fixed epoch cannot be directly explained as a pure sample-selection effect. A higher \(p\) also trains the model more deeply.
	
	The learning rate shows similar progress confounding. With fixed \(p=0.3\) and weight decay, increasing the learning rate from \(3\times 10^{-5}\) to \(3\times 10^{-4}\) decreases the final training loss from about \(0.208\) to about \(0.003\). The test-loss \(p95\) increases from about \(1.108\) to about \(2.689\). Therefore, fixed-epoch results only show that different settings produce different training endpoints. They do not directly show that these settings change the allocation structure at the same training progress. This observation motivates the matched-progress analysis below.
	
	\paragraph{Matched-progress comparison between $p$ and learning rate.}
	We next compare \(p\) and the learning rate under matched progress. For the \(p\) probe, we fix the learning rate at \(10^{-4}\), fix weight decay at \(10^{-2}\), and match different \(p\) values near training loss \(0.5\). The results show that external test-loss quantiles almost collapse. From \(p=0\) to \(p=0.5\), test-loss \(p95\) stays near \(0.90\), and test-loss \(p99\) stays near \(1.0\). This shows that much of the external tail difference caused by \(p\) at a fixed epoch comes from training progress.
	
	However, internal allocation ratios do not collapse. The noise/dense gain ratio increases monotonically from \(1.00\) at \(p=0\) to about \(5.09\) at \(p=0.5\). The noise/dense update ratio increases from about \(0.066\) to about \(0.383\). The weight--bias gradient ratio also increases from about \(17.24\) to about \(22.54\). This means that similar training loss can correspond to similar external loss distributions but different internal coordinate gains and channel update structures.
	
	The learning-rate probe gives the opposite contrast. With fixed \(p=0.3\) and weight decay, we choose matched points near training loss \(0.3\). The three learning rates have test-loss \(p95\) values only between about \(1.01\) and \(1.03\). Their noise/dense gain ratios are also only between about \(2.55\) and \(2.62\), and their noise/dense update ratios are about \(0.23\). Therefore, the learning rate is closer to progress-dominant control. It mainly changes how far the model has progressed, and it does not keep a strong allocation geometry at the same progress. In contrast, \(p\) is a coordinate-gain probe. It changes how the second-moment channel is converted into actual updates through \((v+\epsilon)^{-p}\).
	
	\begin{figure}[t]
		\centering
		\begin{subfigure}[t]{0.48\linewidth}
			\centering
			\begin{tikzpicture}
				\begin{axis}[
					width=\linewidth,
					height=0.72\linewidth,
					xlabel={$p$},
					ylabel={Test loss $p95$},
					grid=both,
					tick label style={font=\scriptsize},
					label style={font=\small}
					]
					\addplot+[mark=*]
					table[x=p,y=TestLossP95,col sep=comma]
					{wilsonfig2matchedpprobet05.csv};
				\end{axis}
			\end{tikzpicture}
			\caption{External loss in the $p$ probe.}
		\end{subfigure}
		\hfill
		\begin{subfigure}[t]{0.48\linewidth}
			\centering
			\begin{tikzpicture}
				\begin{axis}[
					width=\linewidth,
					height=0.72\linewidth,
					xlabel={$p$},
					ylabel={Noise/dense gain ratio},
					grid=both,
					tick label style={font=\scriptsize},
					label style={font=\small}
					]
					\addplot+[mark=*]
					table[x=p,y=GainND,col sep=comma]
					{wilsonfig2matchedpprobet05.csv};
				\end{axis}
			\end{tikzpicture}
			\caption{Internal gain geometry in the $p$ probe.}
		\end{subfigure}
		
		\vspace{0.3em}
		
		\begin{subfigure}[t]{0.48\linewidth}
			\centering
			\begin{tikzpicture}
				\begin{axis}[
					width=\linewidth,
					height=0.72\linewidth,
					xlabel={Learning rate},
					ylabel={Test loss $p95$},
					xmode=log,
					grid=both,
					tick label style={font=\scriptsize},
					label style={font=\small}
					]
					\addplot+[mark=*]
					table[x=lr,y=TestLossP95,col sep=comma]
					{wilsonfig2matchedlrprobet03.csv};
				\end{axis}
			\end{tikzpicture}
			\caption{External loss in the learning-rate probe.}
		\end{subfigure}
		\hfill
		\begin{subfigure}[t]{0.48\linewidth}
			\centering
			\begin{tikzpicture}
				\begin{axis}[
					width=\linewidth,
					height=0.72\linewidth,
					xlabel={Learning rate},
					ylabel={Noise/dense gain ratio},
					xmode=log,
					grid=both,
					tick label style={font=\scriptsize},
					label style={font=\small}
					]
					\addplot+[mark=*]
					table[x=lr,y=GainND,col sep=comma]
					{wilsonfig2matchedlrprobet03.csv};
				\end{axis}
			\end{tikzpicture}
			\caption{Internal gain geometry in the learning-rate probe.}
		\end{subfigure}
		\caption{Matched-progress comparison between $p$ and learning rate. The $p$ probe is matched at train loss $\approx 0.5$. The learning-rate probe is matched at train loss $\approx 0.3$ to keep all three learning-rate settings. External loss differences mostly collapse after matching training progress, but $p$ keeps strong channel-gain and update signatures.}
		\label{fig:wilson_p_lr_matched}
	\end{figure}
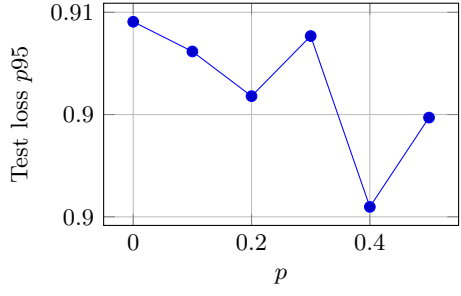
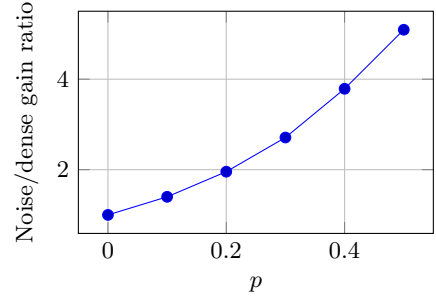
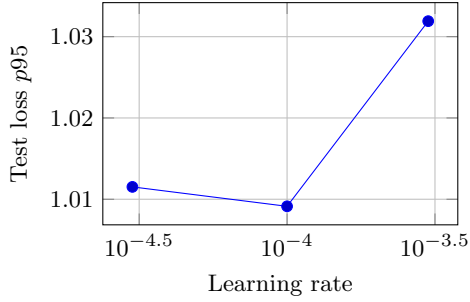
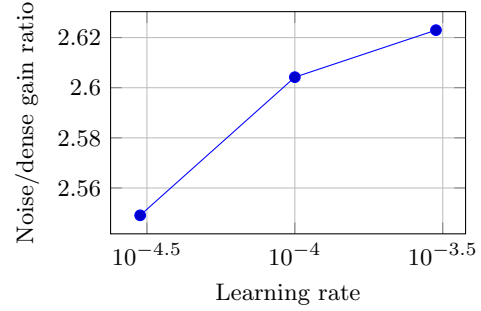
	
	\paragraph{From a single contrast to source signatures.}
	To place all probes under the same protocol, this paper computes a collapse--persistence signature for each training factor. The external range is defined as the log range of test-loss \(p95\) across variants after matched progress. The internal range is defined as the largest log range among \(R_g\), \(R_u/R_g\), the noise/dense gain ratio, and the noise/dense update ratio. This design follows the theoretical hierarchy in Section 3. The external range mainly shows whether the external generalization tail is still different under similar training loss. The internal range shows whether different parameter-path or channel allocation ratios remain under the same training progress.
	
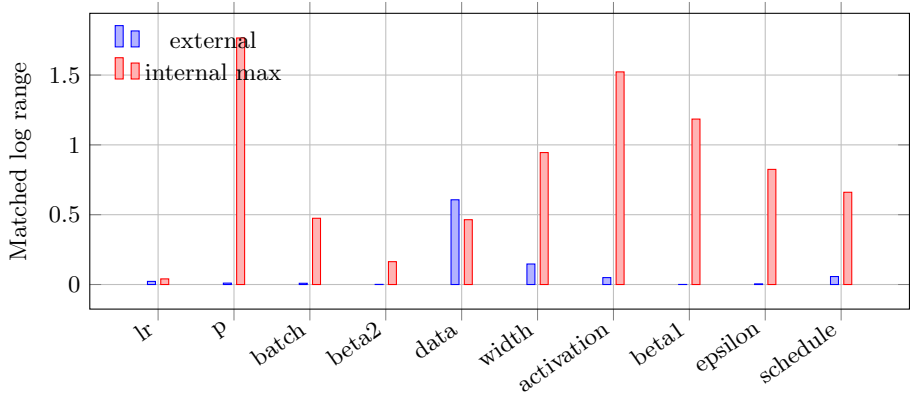
\begin{figure}[t]
		\centering
		\begin{tikzpicture}
			\begin{axis}[
				width=0.95\linewidth,
				height=0.42\linewidth,
				ybar,
				bar width=3pt,
				ylabel={Matched log range},
				symbolic x coords={lr,p,batch,beta2,data,width,activation,beta1,epsilon,schedule},
				xtick=data,
				x tick label style={rotate=35,anchor=east,font=\scriptsize},
				tick label style={font=\scriptsize},
				label style={font=\small},
				legend style={font=\scriptsize,draw=none,fill=none,at={(0.02,0.98)},anchor=north west},
				grid=both
				]
				\addplot table[x=Source,y=ExternalLogRange,col sep=comma]
				{wilsonfig3sourcesignature.csv};
				\addlegendentry{external}
				\addplot table[x=Source,y=InternalMaxLogRange,col sep=comma]
				{wilsonfig3sourcesignature.csv};
				\addlegendentry{internal max}
			\end{axis}
		\end{tikzpicture}
		\caption{Collapse--persistence signatures of representative training-time information-allocation sources. The external range is the residual variation of test-loss $p95$ after matching training loss. The internal range is the largest residual variation among weight--bias and channel-level allocation ratios. The learning rate is closer to a progress-dominant factor. In contrast, $p$, batch size, optimizer memory, gain floor, data statistics, width, activation, and schedule leave different internal allocation signatures.}
		\label{fig:wilson_source_signature}
	\end{figure}

	Figure~\ref{fig:wilson_source_signature} summarizes the signatures of representative sources. The learning rate has an external log range of about \(0.022\) and an internal maximum log range of about \(0.040\). Both are small. Thus, it is closer to a progress-dominant source. In contrast, \(p\) has an external log range of only about \(0.010\), but its internal maximum log range reaches about \(1.77\). This indicates that its main effect remains in the internal channel update geometry after matched progress. Batch size, \(\beta_1\), \(\epsilon\), data, width, and activation also leave different internal signatures. This means that they are not the same scalar effect. They change the training recursion through different entry points.
	
	\paragraph{Stochastic observation and optimizer-state sources.}
	Batch size behaves differently from the learning rate. It is not a single-direction scalar for training progress. It changes the stochastic observation process of mini-batches. For a sparse feature with sample-level occurrence probability \(\pi\), the probability that it is fully absent from a batch is \((1-\pi)^B\). Thus, a larger batch reduces mini-batch observation sparsity. At the same time, mini-batch gradient variance approximately decreases as \(1/B\). The former makes sparse features more likely to be observed continuously. The latter weakens occasional spikes. These two effects jointly change the second-moment EMA. Therefore, batch size can produce non-monotonic or even crossing channel allocation patterns.
	
	The batch probe in Table~\ref{tab:wilson_memory_sources} reflects this point. At the matched point near training loss \(0.3\), increasing batch size from \(32\) to \(512\) keeps test-loss \(p95\) near \(1.01\). However, the noise/dense gain ratio increases from about \(2.40\) to about \(3.19\), while the noise/dense update ratio decreases from about \(0.329\) to about \(0.205\). Gain and update do not move in the same direction. This shows that batch size does not simply amplify or weaken one type of channel. Instead, it changes how sparse channels enter gradients and EMA states.
	
	Optimizer-state probes further separate numerator memory, denominator memory, and gain floor. \(\beta_1\) controls how gradient demand accumulates into a persistent update direction. At matched training loss near \(0.5\), increasing \(\beta_1\) from \(0\) to \(0.9\) leaves test-loss \(p95\) almost unchanged. However, \(R_u/R_g\) decreases from about \(1.26\) to about \(0.69\), and the noise/dense update ratio decreases from about \(0.571\) to about \(0.174\). Thus, the main residual effect of \(\beta_1\) is not external progress. It is the conversion from gradient demand to actual update.
	
	The parameter \(\beta_2\) controls second-moment memory. It determines which historical squared gradients enter the current \(v_t\). Under the same matched-progress condition, increasing \(\beta_2\) from \(0.9\) to \(0.999\) leaves test loss almost unchanged. However, \(R_u/R_g\) decreases from about \(0.813\) to about \(0.690\), and the noise/dense gain ratio increases from about \(2.50\) to about \(2.71\). This shows that \(\beta_2\) changes denominator memory. It first determines the historical statistics in \(v_t\). Then \(p\) converts these statistics into gain.
	
	Finally, \(\epsilon\) is the low-second-moment floor. When \(v_i,v_j\ll\epsilon\), the gain ratio \(((v_j+\epsilon)/(v_i+\epsilon))^p\) is close to \(1\). Therefore, a larger \(\epsilon\) directly limits the amplification of low-\(v\) coordinates. In the experiments, increasing \(\epsilon\) from \(10^{-8}\) to \(10^{-4}\) keeps the test-loss quantiles close. However, the noise/dense gain ratio decreases from about \(2.71\) to about \(1.19\), and the noise/dense update ratio decreases from about \(0.174\) to about \(0.084\). This provides a clean evidence for the gain-floor interpretation.
	
	\begin{table}[t]
		\centering
		\caption{Representative matched-progress rows for stochastic observation and optimizer-state sources.}
		\label{tab:wilson_memory_sources}
		\scriptsize
		\begin{filecontents*}{wilsontabmemorysources.csv}
Probe,Variant,MatchedTrainLoss,TestLossP95,TestLossP99,RgWB,RuDivRg,GainND,UpdateND
batch,B=32,0.292187,1.01812,1.21618,16.1551,0.622676,2.3975,0.328819
batch,B=128,0.299844,1.00913,1.21212,13.682,0.588653,2.6042,0.232575
batch,B=512,0.298024,1.01251,1.20826,12.8442,0.581019,3.18778,0.204612
beta2,beta2=0.9,0.490439,0.903107,1.0163,17.4817,0.812746,2.49733,0.170167
beta2,beta2=0.99,0.492563,0.903793,1.01663,17.5975,0.741167,2.65161,0.173234
beta2,beta2=0.999,0.493462,0.903839,1.01659,17.6515,0.690038,2.71082,0.174486
beta1,beta1=0,0.493538,0.904229,1.0179,17.704,1.26105,2.7087,0.570558
beta1,beta1=0.5,0.493529,0.904167,1.01803,17.6963,0.936,2.70916,0.402285
beta1,beta1=0.9,0.493462,0.903839,1.01659,17.6515,0.690038,2.71082,0.174486
epsilon,eps=1e-08,0.493462,0.903839,1.01659,17.6515,0.690038,2.71082,0.174486
epsilon,eps=1e-06,0.494399,0.90364,1.01627,17.6559,0.693699,2.11055,0.159149
epsilon,eps=0.0001,0.513428,0.899834,1.00913,17.9992,0.8301,1.18858,0.0844511
\end{filecontents*}
\pgfplotstabletypeset[
		col sep=comma,
		columns={Probe,Variant,MatchedTrainLoss,TestLossP95,TestLossP99,RgWB,RuDivRg,GainND,UpdateND},
		columns/Probe/.style={string type,column name={Probe}},
		columns/Variant/.style={string type,column name={Variant}},
		columns/MatchedTrainLoss/.style={column name={Train loss},fixed,precision=3},
		columns/TestLossP95/.style={column name={Test $p95$},fixed,precision=3},
		columns/TestLossP99/.style={column name={Test $p99$},fixed,precision=3},
		columns/RgWB/.style={column name={$R_g$},fixed,precision=2},
		columns/RuDivRg/.style={column name={$R_u/R_g$},fixed,precision=3},
		columns/GainND/.style={column name={Gain N/D},fixed,precision=3},
		columns/UpdateND/.style={column name={Update N/D},fixed,precision=3},
		every head row/.style={before row=\toprule,after row=\midrule},
		every last row/.style={after row=\bottomrule}
		]{wilsontabmemorysources.csv}
	\end{table}
	
	\paragraph{Data-statistical and network-structural sources.}
	The data, width, and activation probes show that information allocation is not only an optimizer-internal phenomenon. Data statistics determine what information can be written. Under matched training loss, the dense useful, sparse useful, and sparse spurious presets have similar empirical risk, but their test distributions remain clearly different. For example, sparse useful has a test accuracy of about \(96.58\%\) and a test-loss \(p95\) of about \(0.670\). Sparse spurious has a test accuracy of about \(51.61\%\) and a test-loss \(p95\) of about \(0.933\). Dense useful has a test accuracy of about \(70.54\%\), but its test-loss tail is higher. This shows that similar empirical risk can be reached through different statistical channels, and these channels do not have the same generalization properties. In other words, the data preset changes available information. It does not only change training speed.
	
	Model width changes the number of input-conditioned weight-like paths. Therefore, it mainly affects gradient-demand allocation. Table~\ref{tab:wilson_structure_sources} shows that at matched training loss near \(0.5\), increasing hidden width from \(16\) to \(256\) raises \(R_g\) from about \(10.40\) to about \(26.75\). This means that a wider model provides more weight-like residual absorption paths. At the same training progress, the residual is more easily projected onto input-dependent paths. This result also shows that weight--bias allocation is not unique to \(p\). Model structure itself can change the ratio by which residual signals enter different parameter paths.
	
	The activation probe shows that nonlinear gating changes Jacobian geometry. For one layer \(h=\phi(u)\), weight and bias gradients contain \(\delta\phi'(u)x\) and \(\delta\phi'(u)\), respectively. Therefore, different activation functions change which samples and features enter gradients through \(\phi'(u)\). At matched training loss near \(0.5\), ReLU, GELU, and LeakyReLU have \(R_g\) values around \(17\)--\(22\), while tanh has an \(R_g\) of about \(80.92\). This does not mean that tanh is better or worse. The key point is that the activation function can strongly change weight--bias gradient routing even when the optimizer is unchanged.
	
	\begin{table}[t]
		\centering
		\caption{Representative matched-progress rows for data-statistical and network-structural sources.}
		\label{tab:wilson_structure_sources}
		\scriptsize
		\begin{filecontents*}{wilsontabstructuresources.csv}
Probe,Variant,MatchedTrainLoss,TestAccPct,TestLossP95,TestLossP99,RgWB,RuDivRg,GainND,UpdateND
data,dense-useful,0.513132,70.5444,1.22969,1.702,22.0065,0.798277,2.55715,0.254152
data,sparse-spurious,0.493462,51.61130000000001,0.932548,1.04641,17.6515,0.690038,2.71082,0.174486
data,sparse-useful,0.495955,96.582,0.670259,0.746667,17.6723,0.688991,2.38003,0.277556
width,16,0.509187,53.186,1.0148,1.13829,10.3961,0.799899,2.76444,0.181382
width,64,0.493462,58.4106,0.903839,1.01659,17.6515,0.690038,2.71082,0.174486
width,256,0.500893,58.2275,0.876152,0.97065,26.7473,0.647846,2.4497,0.180415
activation,gelu,0.491552,58.7158,0.893481,1.00054,18.4521,0.689815,2.81243,0.161397
activation,leakyrelu,0.511507,57.6172,0.904036,1.00955,21.7789,0.705364,2.79977,0.166375
activation,relu,0.493462,58.4106,0.903839,1.01659,17.6515,0.690038,2.71082,0.174486
activation,tanh,0.486186,57.9712,0.938827,1.05832,80.923,1.35624,3.03376,0.148785
\end{filecontents*}
\pgfplotstabletypeset[
		col sep=comma,
		columns={Probe,Variant,MatchedTrainLoss,TestAccPct,TestLossP95,TestLossP99,RgWB,RuDivRg,GainND,UpdateND},
		columns/Probe/.style={string type,column name={Probe}},
		columns/Variant/.style={string type,column name={Variant}},
		columns/MatchedTrainLoss/.style={column name={Train loss},fixed,precision=3},
		columns/TestAccPct/.style={column name={Acc.},fixed,precision=2},
		columns/TestLossP95/.style={column name={Test $p95$},fixed,precision=3},
		columns/TestLossP99/.style={column name={Test $p99$},fixed,precision=3},
		columns/RgWB/.style={column name={$R_g$},fixed,precision=2},
		columns/RuDivRg/.style={column name={$R_u/R_g$},fixed,precision=3},
		columns/GainND/.style={column name={Gain N/D},fixed,precision=3},
		columns/UpdateND/.style={column name={Update N/D},fixed,precision=3},
		every head row/.style={before row=\toprule,after row=\midrule},
		every last row/.style={after row=\bottomrule}
		]{wilsontabstructuresources.csv}
	\end{table}
	
	\paragraph{Trajectory-level control.}
	The schedule probe is not used as main performance evidence. It is used to verify that time-varying choices of \(p\), learning rate, or weight decay change the order of allocation states. Table~\ref{tab:wilson_schedule_summary} shows that the best accuracies of different schedules differ only slightly, but internal ratios and the test tail are not the same. For example, the \(p\)-down schedule gives the highest best accuracy, about \(64.20\%\), but its final test-loss tail is also higher. Combining \(p\)-down with learning-rate-down further lowers final training loss but produces a heavier test tail. This shows that the main role of a schedule is not single-point performance. It is to change the temporal ordering of different allocation states during training. This result is consistent with the training recursion in Section 3. When \(p_t\), \(\eta_t\), or \(\lambda_t\) changes over time, the model changes not only the final hyperparameter values. It also changes the order in which residuals are absorbed by different parameter paths at different training stages.
	
	\begin{table}[t]
		\centering
		\caption{Schedule-level summary. This paper interprets the schedule probe as trajectory-level allocation control, not as a separate accuracy claim.}
		\label{tab:wilson_schedule_summary}
		\scriptsize
		\begin{filecontents*}{wilsonfig7schedulesummary.csv}
ScheduleName,PStart,PEnd,LrStart,LrEnd,WdStart,WdEnd,FinalTrainLoss,FinalTestAcc,TestLossMean,TestLossP95,TestLossP99,RgWB,RuDivRg,GainND,UpdateND,BestTestAcc,BestEpoch,FinalAccPct,BestAccPct
I-fixed-base,0.3,0.3,0.0001,0.0001,0.01,0.01,0.0242550708353519,0.637939453125,0.6609110832214355,1.818952560424805,2.611934185028076,11.784362686606013,0.5490925507401645,2.382761063904563,0.6090192226673978,0.6392822265625,42,63.7939453125,63.92822265625
I-lr-down,0.3,0.3,0.0003,3e-05,0.01,0.01,0.0095367161557078,0.6365966796875,0.6985496878623962,2.216188669204712,3.237318277359009,11.952318179688096,0.5204816690287499,2.2701990131976704,0.6125005334431196,0.6387939453125,13,63.65966796875,63.87939453125
I-p-down,0.5,0.2,0.0001,0.0001,0.01,0.01,0.0062144203111529,0.639892578125,0.7090492248535156,2.314713954925537,3.462444543838501,13.285471081668629,0.4882938121145831,1.694478935486075,0.4491384007866166,0.6419677734375,12,63.9892578125,64.19677734375
I-p-down-lr-down,0.5,0.2,0.0003,3e-05,0.01,0.01,0.0013775558909401,0.6358642578125,0.7991235256195068,2.983700752258301,4.525614738464356,14.05784600025988,0.4390988809015873,1.501039401396301,0.4210179197565149,0.64111328125,3,63.58642578125,64.111328125
I-wd-up,0.3,0.3,0.0001,0.0001,0.0,0.1,0.0253848060965538,0.6383056640625,0.6592252254486084,1.7979274988174438,2.577037811279297,11.744755684191931,0.5491386342117978,2.382882860594602,0.6067794534058613,0.6392822265625,42,63.83056640625,63.92822265625
\end{filecontents*}
\pgfplotstabletypeset[
		col sep=comma,
		columns={ScheduleName,PStart,PEnd,LrStart,LrEnd,WdStart,WdEnd,BestAccPct,BestEpoch,FinalAccPct,TestLossP95,RgWB,RuDivRg,GainND},
		columns/ScheduleName/.style={string type,column name={Schedule}},
		columns/PStart/.style={column name={$p_s$},fixed,precision=2},
		columns/PEnd/.style={column name={$p_e$},fixed,precision=2},
		columns/LrStart/.style={column name={$\eta_s$},sci,sci zerofill,precision=1},
		columns/LrEnd/.style={column name={$\eta_e$},sci,sci zerofill,precision=1},
		columns/WdStart/.style={column name={$\lambda_s$},fixed,precision=3},
		columns/WdEnd/.style={column name={$\lambda_e$},fixed,precision=3},
		columns/BestAccPct/.style={column name={Best acc.},fixed,precision=2},
		columns/BestEpoch/.style={column name={Best ep.}},
		columns/FinalAccPct/.style={column name={Final acc.},fixed,precision=2},
		columns/TestLossP95/.style={column name={Test $p95$},fixed,precision=3},
		columns/RgWB/.style={column name={$R_g$},fixed,precision=2},
		columns/RuDivRg/.style={column name={$R_u/R_g$},fixed,precision=3},
		columns/GainND/.style={column name={Gain N/D},fixed,precision=3},
		every head row/.style={before row=\toprule,after row=\midrule},
		every last row/.style={after row=\bottomrule}
		]{wilsonfig7schedulesummary.csv}
	\end{table}
	
	\paragraph{Summary.}
	The controlled source map shows that training-time implicit bias is not a single scalar effect. The learning rate is closer to progress-dominant control because both its external and internal differences largely collapse after matched progress. The preconditioning exponent \(p\) is a coordinate-gain probe because it keeps strong channel-gain and update signatures after matched progress. Batch size changes mini-batch observation sparsity and gradient burstiness. It therefore acts as a stochastic-observation source. \(\beta_1\), \(\beta_2\), and \(\epsilon\) change first-moment memory, second-moment memory, and the low-\(v\) gain floor, respectively. Data statistics determine which statistical information can be written. Width changes available input-conditioned paths. Activation changes Jacobian gating. Schedule changes the temporal order of these allocation states. Thus, weight--bias and channel-level allocation ratios provide a common diagnostic interface for these heterogeneous sources.
	
	\subsection{Optimization dynamics and sample-level information allocation}
	\label{subsec:optimization_dynamics_sample_allocation}
	
	To test whether the allocation signatures in controlled experiments can be projected to real visual tasks, this paper further records parameter-group statistics and sample-level loss trajectories for fixed \(p\) settings on CIFAR-100 with ResNet-18. This section does not build a full source map. Instead, it focuses on two stable projections in real tasks. The first is the gradient-demand structure between non-bias and bias-like parameters. The second is the final loss distribution and hard-sample identity across sample regions. Specifically, we record the global and layer-wise weight-to-bias gradient norm ratio \(G_w/G_b\). We also save per-sample losses on train and test sets under each fixed \(p\). This allows us to analyze whether different \(p\) values change information allocation at the parameter-path and sample-region levels.
	
	\subsubsection{Weight--bias gradient allocation}
	\label{subsubsec:wb_gradient_allocation}
	
\begin{figure}[h]
		\centering
		
		\begin{subfigure}[t]{0.41\linewidth}
			\centering
			\begin{tikzpicture}
				\begin{axis}[
					width=\linewidth,
					height=0.9\linewidth,
					xlabel={Epoch},
					ylabel={$G_w/G_b$},
					grid=both,
					tick label style={font=\scriptsize},
					label style={font=\small},
					ymode=log,
					xmin=1,
					xmax=200
					]
					\addplot+[color=purple,solid,line width=0.9pt, mark=none]
					table [x=epoch, y=lr001_p_neg0p4, col sep=comma]
					{cifar100_fig4_2a_global_wb_ratio_wide.csv};
					
					\addplot+[color=orange,solid,line width=0.9pt, mark=none]
					table [x=epoch, y=lr001_p_neg0p3, col sep=comma]
					{cifar100_fig4_2a_global_wb_ratio_wide.csv};
					
					\addplot+[color=magenta,solid,line width=0.9pt, mark=none]
					table [x=epoch, y=lr001_p_neg0p2, col sep=comma]
					{cifar100_fig4_2a_global_wb_ratio_wide.csv};
					
					\addplot+[color=red,solid,line width=0.9pt, mark=none]
					table [x=epoch, y=lr001_p_neg0p1, col sep=comma]
					{cifar100_fig4_2a_global_wb_ratio_wide.csv};
					
					\addplot+[color=blue,solid,line width=0.9pt, mark=none]
					table [x=epoch, y=lr001_p_pos0p0, col sep=comma]
					{cifar100_fig4_2a_global_wb_ratio_wide.csv};
					
					\addplot+[color=cyan,solid,line width=0.9pt, mark=none]
					table [x=epoch, y=lr001_p_pos0p1, col sep=comma]
					{cifar100_fig4_2a_global_wb_ratio_wide.csv};
					
					\addplot+[color=green,solid,line width=0.9pt, mark=none]
					table [x=epoch, y=lr001_p_pos0p2, col sep=comma]
					{cifar100_fig4_2a_global_wb_ratio_wide.csv};
					
					\addplot+[color=violet,solid,line width=0.9pt, mark=none]
					table [x=epoch, y=lr001_p_pos0p3, col sep=comma]
					{cifar100_fig4_2a_global_wb_ratio_wide.csv};
				\end{axis}
			\end{tikzpicture}
			\caption{$\eta=0.01$}
			\label{fig:global_wb_ratio_lr001}
		\end{subfigure}
		\hspace{0.035\linewidth}
		\begin{subfigure}[t]{0.41\linewidth}
			\centering
			\begin{tikzpicture}
				\begin{axis}[
					width=\linewidth,
					height=0.9\linewidth,
					xlabel={Epoch},
					grid=both,
					tick label style={font=\scriptsize},
					label style={font=\small},
					ymode=log,
					xmin=1,
					xmax=200,
					clip=false,
					legend style={
						font=\scriptsize,
						at={(1.1,1.00)},
						anchor=north west,
						draw=none,
						fill=none,
						row sep=1pt
					}
					]
					\addplot+[color=purple,solid,line width=0.9pt, mark=none]
					table [x=epoch, y=lr01_p_neg0p4, col sep=comma]
					{cifar100_fig4_2a_global_wb_ratio_wide.csv};
					\addlegendentry{$p=-0.4$}
					
					\addplot+[color=orange,solid,line width=0.9pt, mark=none]
					table [x=epoch, y=lr01_p_neg0p3, col sep=comma]
					{cifar100_fig4_2a_global_wb_ratio_wide.csv};
					\addlegendentry{$p=-0.3$}
					
					\addplot+[color=magenta,solid,line width=0.9pt, mark=none]
					table [x=epoch, y=lr01_p_neg0p2, col sep=comma]
					{cifar100_fig4_2a_global_wb_ratio_wide.csv};
					\addlegendentry{$p=-0.2$}
					
					\addplot+[color=red,solid,line width=0.9pt, mark=none]
					table [x=epoch, y=lr01_p_neg0p1, col sep=comma]
					{cifar100_fig4_2a_global_wb_ratio_wide.csv};
					\addlegendentry{$p=-0.1$}
					
					\addplot+[color=blue,solid,line width=0.9pt, mark=none]
					table [x=epoch, y=lr01_p_pos0p0, col sep=comma]
					{cifar100_fig4_2a_global_wb_ratio_wide.csv};
					\addlegendentry{$p=0.0$}
					
					\addplot+[color=cyan,solid,line width=0.9pt, mark=none]
					table [x=epoch, y=lr01_p_pos0p1, col sep=comma]
					{cifar100_fig4_2a_global_wb_ratio_wide.csv};
					\addlegendentry{$p=0.1$}
					
					\addplot+[color=green,solid,line width=0.9pt, mark=none]
					table [x=epoch, y=lr01_p_pos0p2, col sep=comma]
					{cifar100_fig4_2a_global_wb_ratio_wide.csv};
					\addlegendentry{$p=0.2$}
					
					\addplot+[color=violet,solid,line width=0.9pt, mark=none]
					table [x=epoch, y=lr01_p_pos0p3, col sep=comma]
					{cifar100_fig4_2a_global_wb_ratio_wide.csv};
					\addlegendentry{$p=0.3$}
				\end{axis}
			\end{tikzpicture}
			\caption{$\eta=0.1$}
			\label{fig:global_wb_ratio_lr01}
		\end{subfigure}
		
		\caption{Global weight-to-bias gradient norm ratio during training.
			The ratio $G_w/G_b$ is computed from epoch-mean gradient norms of non-bias and bias parameters.
			Across both learning-rate settings, larger $p$ is associated with a lower global weight-to-bias gradient ratio, indicating that changing $p$ leads to distinct parameter-path gradient-demand structures through training trajectory feedback.}
		\label{fig:global_wb_ratio}
	\end{figure}
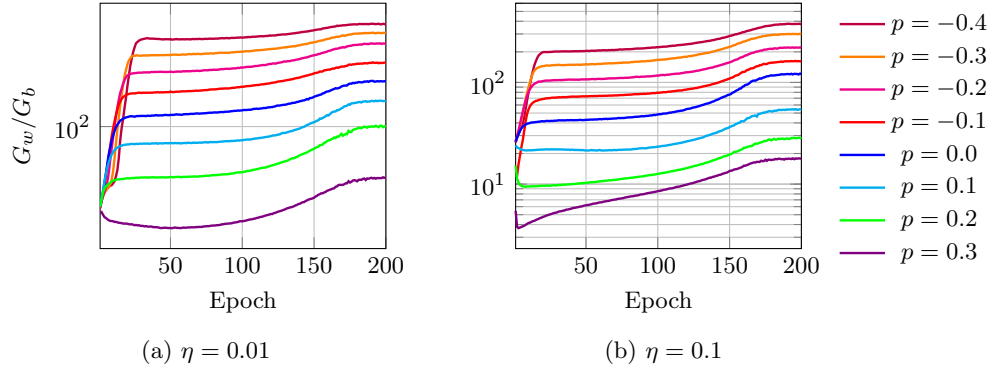

	Figure~\ref{fig:global_wb_ratio} shows the global weight-to-bias gradient norm ratio under different fixed \(p\) values. The result shows a clear monotonic trend. Under learning rate \(\eta=0.01\), the global ratio at the final epoch decreases from about \(565\) at \(p=-0.4\) to about \(42\) at \(p=0.3\). Under the larger learning rate \(\eta=0.1\), the same trend remains. The final ratio decreases from about \(374\) at \(p=-0.4\) to about \(18\) at \(p=0.3\). This shows that different \(p\) values correspond to different gradient-demand structures for non-bias and bias parameters.
	
	This phenomenon has two implications. First, in real networks, changing \(p\) does not only change final accuracy. It is also associated with different parameter-path gradient-demand structures. The exponent \(p\) directly acts on EMA preconditioning gain in the current step. However, the gradient itself is affected by feedback from later parameter positions, residual structures, and activation states. Therefore, \(G_w/G_b\) should be understood as a training-trajectory-level allocation projection, not as a direct preconditioning factor in the current step. Second, this trend is observed under both learning-rate settings. It is not an accidental phenomenon under a single learning rate. Still, the learning rate modulates the strength of the effect. At the same \(p\), a larger learning rate usually corresponds to a lower global \(G_w/G_b\). This is consistent with the earlier analysis of progress--allocation coupling.
	
	The layer-wise statistics in the appendix further show that the global trend is not produced by all layers behaving identically. Many convolutional layers and downsample layers follow the global trend. As \(p\) increases, their weight-to-bias gradient ratios decrease. In contrast, some normalization layers respond weakly or may even move in a different direction. This means that the global \(G_w/G_b\) is the result of superposed multi-layer differential paths. Layer-wise differences may be jointly related to input statistics, normalization states, residual structure, and local gradient propagation. Therefore, implicit bias in real networks is better understood as a mixed projection of several allocation effects rather than a single layer-level rule.
	
	\subsubsection{Sample-level loss distribution}
	\label{subsubsec:sample_loss_distribution}
	
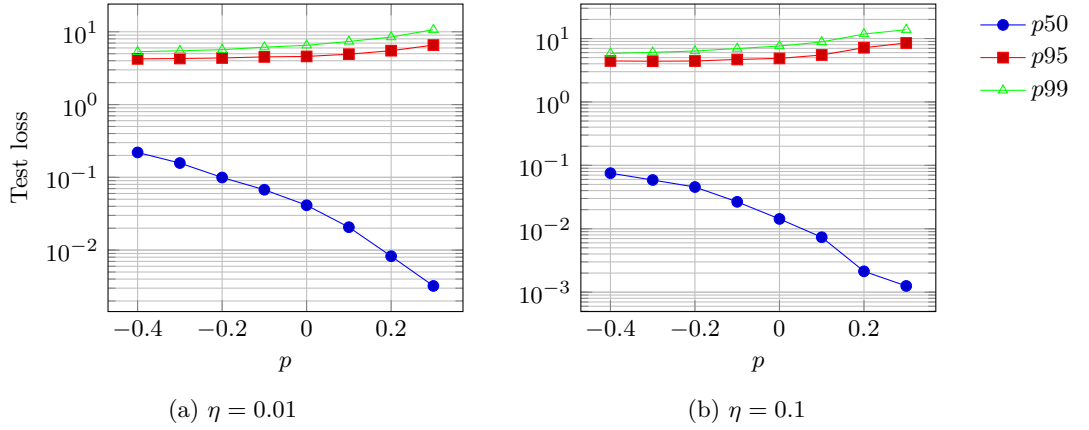
\begin{figure}[t]
		\centering
		
		\begin{subfigure}[h]{0.48\linewidth}
			\centering
			\begin{tikzpicture}
				\begin{axis}[
					width=\linewidth,
					height=0.9\linewidth,
					xlabel={$p$},
					ylabel={Test loss},
					grid=both,
					ymode=log,
					tick label style={font=\scriptsize},
					label style={font=\small}
					]
					\addplot+[color=blue,mark=*] table [x=p, y=lr001_loss_p50, col sep=comma] {cifar100_fig4_2b_test_loss_quantiles_wide.csv};
					
					\addplot+[color=red,mark=square*] table [x=p, y=lr001_loss_p95, col sep=comma] {cifar100_fig4_2b_test_loss_quantiles_wide.csv};
					
					\addplot+[color=green,mark=triangle] table [x=p, y=lr001_loss_p99, col sep=comma] {cifar100_fig4_2b_test_loss_quantiles_wide.csv};
				\end{axis}
			\end{tikzpicture}
			\caption{$\eta=0.01$}
			\label{fig:test_loss_quantiles_lr001}
		\end{subfigure}
		\hfill
		\begin{subfigure}[h]{0.48\linewidth}
			\centering
			\begin{tikzpicture}
				\begin{axis}[
					width=\linewidth,
					height=0.9\linewidth,
					xlabel={$p$},
					grid=both,
					ymode=log,
					legend style={
						font=\scriptsize,
						at={(1.1,1.00)},
						anchor=north west,
						draw=none,
						fill=none,
						row sep=1pt
					},
					tick label style={font=\scriptsize},
					label style={font=\small}
					]
					\addplot+[color=blue,mark=*] table [x=p, y=lr01_loss_p50, col sep=comma] {cifar100_fig4_2b_test_loss_quantiles_wide.csv};
					\addlegendentry{$p50$}
					
					\addplot+[color=red,mark=square*] table [x=p, y=lr01_loss_p95, col sep=comma] {cifar100_fig4_2b_test_loss_quantiles_wide.csv};
					\addlegendentry{$p95$}
					
					\addplot+[color=green,mark=triangle] table [x=p, y=lr01_loss_p99, col sep=comma] {cifar100_fig4_2b_test_loss_quantiles_wide.csv};
					\addlegendentry{$p99$}
				\end{axis}
			\end{tikzpicture}
			\caption{$\eta=0.1$}
			\label{fig:test_loss_quantiles_lr01}
		\end{subfigure}
		
		\caption{Test loss quantiles under different values of $p$.
			Larger $p$ consistently reduces the median loss, but increases the high-loss tail.
			This median--tail trade-off is more pronounced under the larger learning rate.}
		\label{fig:sample_loss_distribution}
	\end{figure}

	Next, this paper analyzes how parameter-allocation dynamics are reflected in sample-level loss distributions. Figure~\ref{fig:sample_loss_distribution} shows changes in clean-train and test-loss quantiles under different \(p\) values. The result shows a clear median--tail trade-off. As \(p\) increases, the test median loss decreases monotonically. This means that a larger \(p\) makes the model fit most samples or core samples with higher confidence. However, the high-loss tail becomes worse. Under \(\eta=0.01\), the test median loss decreases from about \(0.220\) at \(p=-0.4\) to about \(0.003\) at \(p=0.3\), but the test \(p95\) loss increases from about \(4.25\) to about \(6.56\). At the same time, the proportion of test samples with loss larger than \(5\) increases from about \(1.78\%\) to about \(8.48\%\).
	
	Under the larger learning rate \(\eta=0.1\), this phenomenon is more pronounced. As \(p\) increases, the test median loss continues to decrease, but the tail loss becomes worse more strongly. When \(p=0.3\), the test \(p95\) loss reaches about \(8.51\), and the proportion of test samples with loss larger than \(5\) increases to about \(11.07\%\). This indicates that positive \(p\) values more strongly reduce the loss of low-loss or core samples. Under a larger learning rate, this concentrated fitting further amplifies the risk in the high-loss tail.
	
	\begin{table}[t]
		\centering
		\caption{Final test loss distribution summary.}
		\label{tab:test_loss_distribution_summary}
		\scriptsize
		\begin{filecontents*}{cifar100_tab4_2_test_loss_summary.csv}
lr,p,acc_percent,loss_mean,loss_std,loss_p50,loss_p95,loss_p99,frac_loss_gt_5_percent
0.01,-0.4,77.57999897003174,0.9872718421936036,1.4168818012422306,0.2198486328125,4.24609375,5.3203125,1.78
0.01,-0.3,78.53000164031982,0.9212148333191872,1.4242721296090144,0.15716552734375,4.297070312499997,5.484453125000002,2.1399999999999997
0.01,-0.2,78.40999960899353,0.8887717651367187,1.461565560933611,0.099517822265625,4.367382812499997,5.679765625000002,2.56
0.01,-0.1,78.1499981880188,0.8757429504036903,1.5314674624298403,0.06756591796875,4.519726562499997,6.1484375,3.2399999999999998
0.01,0.0,78.65999937057495,0.859466770541668,1.578037619670687,0.04132080078125,4.586132812499997,6.523476562500001,3.6799999999999997
0.01,0.1,77.86999940872192,0.8888077940583229,1.7234646223651282,0.0206527709960937,4.922070312499997,7.367304687500003,4.78
0.01,0.2,77.7899980545044,0.9471313146471976,1.9555438618535996,0.0082435607910156,5.5,8.461562500000014,6.32
0.01,0.3,77.0900011062622,1.109957288336754,2.367757158696816,0.003218650817871,6.563281249999989,10.687812500000009,8.48
0.1,-0.4,78.15999984741211,0.8723625084996224,1.4869827363130022,0.07525634765625,4.429882812499997,5.824531250000007,2.81
0.1,-0.3,78.24000120162964,0.8619079359412193,1.5118759101344352,0.0587310791015625,4.390820312499997,6.062578125000002,2.85
0.1,-0.2,78.4600019454956,0.8408274319052697,1.529272995687226,0.0455169677734375,4.421875,6.355507812500001,3.2399999999999998
0.1,-0.1,78.33999991416931,0.8600700465202331,1.6458554756359007,0.0265884399414062,4.676171874999994,6.949296875000002,4.06
0.1,0.0,78.6899983882904,0.863161622440815,1.7600151795391272,0.0142898559570312,4.859570312499997,7.613320312500001,4.66
0.1,0.1,77.28000283241272,0.973225958263874,2.005599416543241,0.0073509216308593,5.500195312499997,8.836015625000002,6.21
0.1,0.2,76.23000144958496,1.1954879014015198,2.598597949199452,0.0021314620971679,7.164843749999989,11.773828125000009,9.34
0.1,0.3,75.27999877929688,1.4012551540732383,3.090723306002947,0.0012531280517578,8.508593749999989,13.789296875000003,11.07
\end{filecontents*}
\pgfplotstabletypeset[
		col sep=comma,
		columns={lr,p,acc_percent,loss_mean,loss_std,loss_p50,loss_p95,loss_p99,frac_loss_gt_5_percent},
		columns/lr/.style={
			column name={$\eta$},
			fixed,
			precision=2
		},
		columns/p/.style={
			column name={$p$},
			fixed,
			precision=1
		},
		columns/acc_percent/.style={
			column name={Acc.},
			fixed,
			precision=2
		},
		columns/loss_mean/.style={
			column name={Mean},
			fixed,
			precision=3
		},
		columns/loss_std/.style={
			column name={Std.},
			fixed,
			precision=3
		},
		columns/loss_p50/.style={
			column name={$p50$},
			fixed,
			precision=3
		},
		columns/loss_p95/.style={
			column name={$p95$},
			fixed,
			precision=3
		},
		columns/loss_p99/.style={
			column name={$p99$},
			fixed,
			precision=3
		},
		columns/frac_loss_gt_5_percent/.style={
			column name={$L>5$},
			fixed,
			precision=2
		},
		every head row/.style={
			before row=\toprule,
			after row=\midrule
		},
		every last row/.style={
			after row=\bottomrule
		}
		]{cifar100_tab4_2_test_loss_summary.csv}
	\end{table}
	
	The clean-train distribution shows a complementary trend. A larger \(p\) consistently gives lower training-set loss. For example, under \(\eta=0.01\), the clean-train mean loss decreases from about \(0.092\) at \(p=-0.4\) to about \(0.0007\) at \(p=0.3\). Therefore, the worse test tail is not caused by insufficient training-set optimization. Instead, it shows that the fitting result is redistributed across sample regions. A larger \(p\) more strongly reduces the loss of most training samples, but this reduction does not transfer uniformly to all test samples. This produces a clearer median--tail trade-off.
	
	These results do not support the monotonic claim that a smaller \(p\) necessarily gives better generalization. An overly small \(p\) can reduce tail amplification, but it can also weaken training-set fitting. An overly large \(p\) can reduce the loss of most training samples more aggressively, but it can produce a heavier test tail. Better performance often appears in an intermediate allocation region, where fitting of most samples and tail risk are more balanced.
	
	\subsubsection{Hard-sample migration}
	\label{subsubsec:hard_sample_migration}
	
	To test whether the change in the loss tail is only a rescaling of the same difficult samples, or whether the sample priority truly changes, this paper further compares top-loss sample sets under different \(p\) values. We define hard samples as the top \(10\%\) samples ranked by final loss. We compute the Jaccard overlap between different hard-sample sets.
	
\begin{figure}[t]
		\centering
		
		\begin{subfigure}[t]{0.41\linewidth}
			\centering
			\begin{tikzpicture}
				\begin{axis}[
					width=\linewidth,
					height=\linewidth,
					xlabel={$p$},
					ylabel={$p$},
					xmin=-0.5,
					xmax=7.5,
					ymin=-0.5,
					ymax=7.5,
					xtick={0,1,2,3,4,5,6,7},
					ytick={0,1,2,3,4,5,6,7},
					xticklabels={$-0.4$,$-0.3$,$-0.2$,$-0.1$,$0.0$,$0.1$,$0.2$,$0.3$},
					yticklabels={$-0.4$,$-0.3$,$-0.2$,$-0.1$,$0.0$,$0.1$,$0.2$,$0.3$},
					tick label style={font=\scriptsize},
					label style={font=\small},
					axis on top,
					enlargelimits=false,
					point meta min=0,
					point meta max=1,
					colormap/viridis
					]
					\addplot[
					matrix plot*,
					mesh/cols=8,
					point meta=explicit
					]
					table [
					x=x,
					y=y,
					meta=jaccard,
					col sep=comma
					] {cifar100_fig4_2c_hard_overlap_train_eval_lr001_matrix.csv};
				\end{axis}
			\end{tikzpicture}
			\caption{$\eta=0.01$}
			\label{fig:hard_overlap_lr001}
		\end{subfigure}
		\hspace{0.035\linewidth}
		\begin{subfigure}[t]{0.41\linewidth}
			\centering
			\begin{tikzpicture}
				\begin{axis}[
					width=\linewidth,
					height=\linewidth,
					xlabel={$p$},
					xmin=-0.5,
					xmax=7.5,
					ymin=-0.5,
					ymax=7.5,
					xtick={0,1,2,3,4,5,6,7},
					ytick={0,1,2,3,4,5,6,7},
					xticklabels={$-0.4$,$-0.3$,$-0.2$,$-0.1$,$0.0$,$0.1$,$0.2$,$0.3$},
					yticklabels={},
					tick label style={font=\scriptsize},
					label style={font=\small},
					axis on top,
					enlargelimits=false,
					point meta min=0,
					point meta max=1,
					colormap/viridis,
					colorbar,
					colorbar style={
						ylabel={Jaccard},
						tick label style={font=\scriptsize},
						label style={font=\small}
					}
					]
					\addplot[
					matrix plot*,
					mesh/cols=8,
					point meta=explicit
					]
					table [
					x=x,
					y=y,
					meta=jaccard,
					col sep=comma
					] {cifar100_fig4_2c_hard_overlap_train_eval_lr01_matrix.csv};
				\end{axis}
			\end{tikzpicture}
			\caption{$\eta=0.1$}
			\label{fig:hard_overlap_lr01}
		\end{subfigure}
		
		\caption{Hard-sample overlap across different values of $p$.
			Hard samples are defined as the top $10\%$ samples ranked by final clean-train loss.
			Each cell reports the Jaccard overlap between two hard-sample sets.
			Lower off-diagonal overlap indicates stronger hard-sample migration, showing that changing $p$ modifies which samples remain in the high-loss tail.}
		\label{fig:hard_sample_overlap}
	\end{figure}
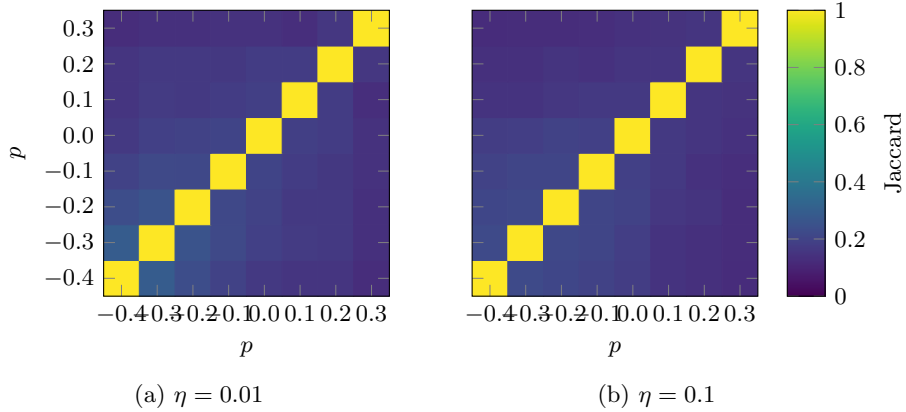

	Figure~\ref{fig:hard_sample_overlap} shows that hard-sample sets on the clean training set are neither random nor fixed. Under \(\eta=0.01\), the Jaccard overlap between adjacent \(p\) values is about \(0.16\)--\(0.29\), while the overlap between \(p=-0.4\) and \(p=0.3\) decreases to about \(0.125\). Similar results are observed under \(\eta=0.1\). The low overlap on the training set shows that changing \(p\) changes which samples remain in the high-loss tail during training. Therefore, \(p\) does not only scale the loss distribution. It changes sample-level optimization priority.
	
	This observation is key evidence for sample-region allocation. If changing \(p\) only rescales loss globally, then the identity of hard samples should remain mostly stable. However, the experiments show that hard samples on the clean training set migrate clearly as \(p\) changes. This means that different preconditioning exponents allocate fitting capacity to different sample sets.
	
	\subsubsection{Interaction with learning-rate schedules}
	\label{subsubsec:lr_decay_allocation}
	
	Although this section mainly compares fixed \(p\), all runs use the same cosine learning-rate schedule. Therefore, changes in late-stage \(G_w/G_b\) also provide evidence for progress--allocation coupling. In most runs, the gradient norms of weight and bias-like parameters both decrease as training proceeds, but their rates of decrease are not identical. Thus, the global \(G_w/G_b\) often changes systematically in late training.
	
	This phenomenon shows that the weight--bias gradient structure is not an isolated result of \(p\). It is jointly determined by the preconditioning exponent, the learning-rate trajectory, the residual state, and the local differential structure of the network. This is consistent with the controlled Wilson-style experiments. The learning rate mainly controls the speed of training progress and changes later gradient demand through trajectory feedback. The exponent \(p\) more directly changes coordinate gain under EMA preconditioning. They are coupled during multi-step training. Therefore, \(G_w/G_b\) in real tasks should be understood as an observable projection after progress and allocation interact.
	
	For this reason, this section does not interpret the learning-rate schedule as a separate bias-calibration mechanism. Instead, it treats it as an expression of progress--allocation coupling in real training. The same \(p\) can produce different parameter-path statistics at different learning-rate stages. The same learning-rate schedule can also lead to different sample loss distributions and hard-sample structures under different \(p\) values.
	
	\subsubsection{Summary}
	\label{subsubsec:optimization_dynamics_summary}
	
	The results in this section support three conclusions.
	
	First, in real training on CIFAR-100 with ResNet-18, fixed \(p\) values correspond to different global weight-to-bias gradient norm ratios. This trend is stable under both learning-rate settings. It shows that the allocation effect observed in controlled experiments can be projected to parameter-path statistics in real visual networks.
	
	Second, changing \(p\) reshapes the sample-level loss distribution. A larger \(p\) reduces the median loss but increases the high-loss tail. This effect is stronger under the larger learning rate. This indicates that different preconditioning exponents do not only change training-set fitting strength. They also change the allocation relation between majority samples and tail samples.
	
	Third, hard-sample overlap analysis shows that different \(p\) values produce different hard-sample sets, especially on the clean training set. This means that \(p\) controls not only the global optimization magnitude, but also how fitting capacity is distributed across sample sets.
	
	Taken together, the CIFAR-100 and ResNet-18 results provide real-task evidence for the information-allocation view. The exponent \(p\) is not a simple scalar interpolation between SGD-like and Adam-like behavior. It changes coordinate gain through EMA preconditioning. In real networks, this appears as joint changes in parameter-path statistics, sample loss distributions, and hard-sample identities.
	
	\subsection{Dynamic information allocation on facial expression recognition}
	\label{subsec:dynamic_allocation_fer}
	
	To test the practical value of allocation-aware scheduling, this paper further evaluates dynamic \(p\) schedules on facial expression recognition tasks based on PosterV2. Unlike the fixed-\(p\) mechanism analysis above, this section does not try to decompose all training factors. It asks a simpler question. In real FER tasks, can adjusting the strength of EMA preconditioning across training stages improve convergence behavior or final performance?
	
	\begin{figure}[h]
		\centering
		
		\begin{subfigure}[h]{0.4\linewidth}
			\centering
			\begin{tikzpicture}
				\begin{axis}[
					width=\linewidth,
					height=0.9\linewidth,
					xlabel={Epoch},
					ylabel={Accuracy (\%)},
					xmin=0,
					xmax=200,
					ymin=90,
					ymax=93,
					grid=both,
					tick label style={font=\scriptsize},
					label style={font=\small},
					legend style={
						font=\scriptsize,
						at={(1.05,1.0)},
						anchor=north west,
						draw=none,
						fill=none
					}
					]
					\addplot+[color=blue,mark=none,line width=0.9pt]
					table[x=step,y=rafdb_origin,col sep=comma]
					{rafdb_acc_wide.csv};
					\addlegendentry{$p=0.5$}
					\addplot+[color=red,mark=none,line width=0.9pt]
					table[x=step,y=rafdb_dynamic,col sep=comma]
					{rafdb_acc_wide.csv};
					\addlegendentry{$p:0.45\rightarrow0.35$}
				\end{axis}
			\end{tikzpicture}
			\caption{RAFDB}
		\end{subfigure}
		\hfill
		\begin{subfigure}[h]{0.4\linewidth}
			\centering
			\begin{tikzpicture}
				\begin{axis}[
					width=\linewidth,
					height=0.9\linewidth,
					xlabel={Iteration},
					yticklabel style={},
					xmin=0,
					xmax=62000,
					ymin=66.5,
					ymax=67.7,
					restrict y to domain=66.5:67.7,
					grid=both,
					tick label style={font=\scriptsize},
					label style={font=\small},
					clip=false,
					legend style={
						font=\scriptsize,
						at={(1.05,1.0)},
						anchor=north west,
						draw=none,
						fill=none
					}
					]
					\addplot+[color=blue,mark=none,line width=0.9pt]
					table[x=step,y=affect_fixed,col sep=comma]
					{affectnet7_acc_wide.csv};
					\addlegendentry{$p=0.5$}
					
					\addplot+[color=red,mark=none,line width=0.9pt]
					table[x=step,y=affect_dynamic_06_05,col sep=comma]
					{affectnet7_acc_wide.csv};
					\addlegendentry{$p:0.6\rightarrow0.5$}
					
					\addplot+[color=green,mark=none,line width=0.9pt]
					table[x=step,y=affect_dynamic_055_0,col sep=comma]
					{affectnet7_acc_wide.csv};
					\addlegendentry{$p:0.55\rightarrow0.4$}
					
				\end{axis}
			\end{tikzpicture}
			\caption{AffectNet7}
		\end{subfigure}
		
		\caption{Dynamic $p$ scheduling on FER benchmarks. The allocation-aware schedules achieve slightly higher best accuracy and, on AffectNet7, faster convergence to the best-performing region.}
		\label{fig:fer_dynamic_acc}
	\end{figure}
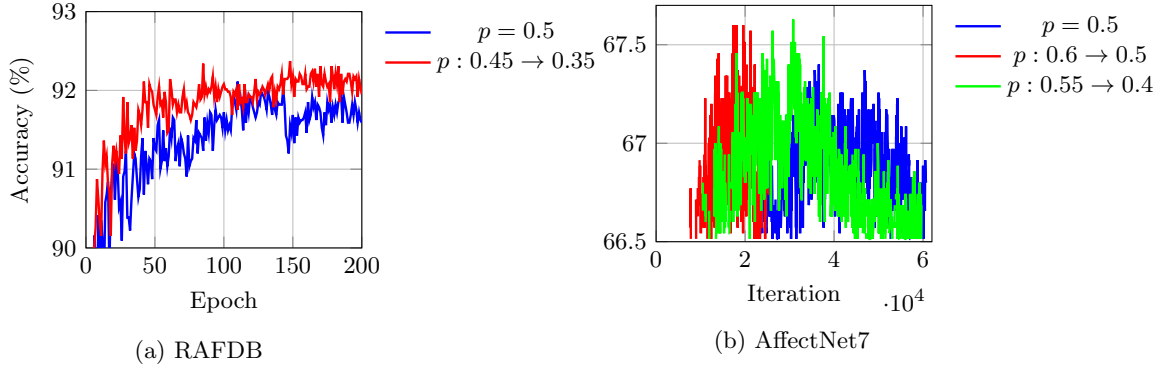
	
	First, we conduct experiments on RAFDB. As shown in Fig.~\ref{fig:fer_dynamic_acc}, the original training strategy with \(p=0.5\) reaches a best accuracy of \(92.21\%\). With the dynamic information-allocation strategy, the accuracy increases to \(92.37\%\). In this experiment, \(p\) gradually decreases from \(0.45\) to \(0.35\). RAFDB is relatively small. The calibration between pretrained representations and the task classifier head may be sensitive to the late-stage update state. The result shows that properly reducing \(p\) during training can change the late-stage update state and produce a small but stable performance gain.
	
	On the more challenging AffectNet7 dataset, dynamic information allocation shows a clearer advantage. This dataset has larger intra-class variation and more complex expression patterns. The model must balance main representation learning and difficult-sample optimization. As shown in Fig.~\ref{fig:fer_dynamic_acc}, the baseline with fixed \(p=0.5\) reaches a best accuracy of \(67.49\%\). In contrast, the dynamic schedule that gradually decreases \(p\) from \(0.6\) to \(0.5\) improves the best accuracy to \(67.61\%\). The dynamic strategy between \(0.6\) and \(0.5\) also shows faster convergence.
	
	In particular, compared with fixed \(p=0.5\), the dynamic \(p\) strategy reaches a higher best accuracy using about half the training steps. This suggests that dynamic information allocation can improve not only final performance, but also optimization efficiency. This may be related to the larger data scale and more complex intra-class variation of AffectNet7. A larger \(p\) in early training may strengthen coordinate selectivity under EMA preconditioning and reduce the loss of main sample regions faster. As training proceeds, decreasing \(p\) may weaken excessive coordinate selectivity and improve late-stage calibration and tail-sample handling. This explanation is consistent with the median--tail trade-off observed above. However, its exact mechanism still requires finer sample-level and parameter-group diagnostics.
	
	These results show that allocation-aware dynamic scheduling has practical potential in FER tasks. Different data scales, pretrained states, and task difficulties may require different information-allocation states. Therefore, a fixed \(p\) may not always fit both early representation adjustment and late sample calibration. Dynamic \(p\) provides a simple way to adjust the strength of EMA preconditioning across training stages without changing the model structure or the loss function. It can improve convergence or performance in real tasks.
	
	\section{Conclusion}\label{sec7}
	
	This paper proposes an information-allocation view of training-time implicit bias. The main idea is that optimization does not only move a model toward a final solution. It also decides how error signals are written into parameter paths, coordinate channels, optimizer states, and sample regions during training. This view provides a common diagnostic layer for heterogeneous factors such as learning rate, preconditioning exponent, batch size, optimizer memory, weight decay, data statistics, model width, activation function, and schedule.
	
	The controlled experiments show that different sources leave different collapse--persistence signatures. The learning rate mainly acts as a progress-dominant source. Its external and internal differences largely collapse after matching training loss. In contrast, the exponent \(p\) keeps strong channel-gain and update signatures after progress matching. Batch size, optimizer memory, gain floors, data statistics, network width, activation functions, and schedules also produce distinct internal allocation patterns.
	
	Experiments on CIFAR-100 with ResNet-18 show that changing \(p\) changes the weight--bias gradient structure, sample-wise loss quantiles, and hard-sample identity. A larger \(p\) tends to reduce the median loss, but it can increase the high-loss tail. This gives a concrete median--tail trade-off. Experiments on facial expression recognition further show that dynamic \(p\) schedules can improve convergence and final accuracy. These results suggest that allocation-aware optimization is not only a diagnostic tool. It can also guide practical training strategies.
	
	This work has limitations. The weight--bias decomposition is a minimal projection, not a complete description of all network dynamics. Other decompositions, such as layer-wise, module-wise, frequency-domain, and sample-group decompositions, should be studied further. The dynamic schedules used here are also simple. Future work should develop adaptive methods that separately control global progress and internal allocation. Such methods may provide more stable training and better generalization across tasks, architectures, and data distributions.
	
	\bibliography{sn-bibliography}
	
\end{document}